\documentclass[10pt,twocolumn,letterpaper]{article}

\usepackage{cvpr}
\usepackage{times}
\usepackage{epsfig}
\usepackage{graphicx}
\usepackage{amsmath}
\usepackage{amssymb}
\usepackage{ownstyles}
\usepackage{textpos}

\newif\iftodos

\todostrue

\usepackage[usenames,dvipsnames]{xcolor} 
\iftodos

\newcommand{\todo}[1]{\textcolor{red}{[#1]}} 
\newcommand{\done}[1]{\textcolor{Emerald}{[#1]}} 
\newcommand{\comment}[1]{\textcolor{blue}{[#1]}} 
\newcommand{\jmccomment}[1]{\textcolor{magenta}{[#1]}} 
\newcommand{\jbycomment}[1]{\textcolor{Orange}{[JBY: #1]}} 
\newcommand{\lowpriority}[1]{\textcolor{green}{[#1]}} 
\else
\newcommand{\todo}[1]{} 
\newcommand{\done}[1]{} 
\newcommand{\comment}[1]{} 
\newcommand{\jmccomment}[1]{} 
\newcommand{\jbycomment}[1]{} 
\newcommand{\lowpriority}[1]{} 
\fi

\usepackage[pagebackref=true,breaklinks=true,letterpaper=true,colorlinks,bookmarks=false]{hyperref}

\cvprfinalcopy 

\def\cvprPaperID{1307} 

\ifcvprfinal\fi
\begin{document}

\title{Deep Neural Networks are Easily Fooled:\\High Confidence Predictions for Unrecognizable Images}

\author{Anh Nguyen\\
University of Wyoming\\
{\tt\small anguyen8@uwyo.edu}
\and
Jason Yosinski\\
Cornell University\\
{\tt\small yosinski@cs.cornell.edu}
\and
Jeff Clune\\
University of Wyoming\\
{\tt\small jeffclune@uwyo.edu}
}

\maketitle

\begin{textblock*}{2\columnwidth}(.0\textwidth,-7.0cm)
	Full Citation: Nguyen A, Yosinski J, Clune J. \textit{Deep Neural Networks are Easily Fooled: High Confidence Predictions for Unrecognizable Images}. In Computer Vision and Pattern Recognition (CVPR '15), IEEE, 2015.
\end{textblock*}
\vspace{-0.05in} 

\begin{abstract}

Deep neural networks (DNNs) have recently been achieving state-of-the-art performance on a variety of pattern-recognition tasks, most notably visual classification problems. Given that DNNs are now able to classify objects in images with near-human-level performance, questions naturally arise as to what differences remain between computer and human vision. A recent study \cite{szegedy2013intriguing} revealed that changing an image (e.g. of a lion) in a way imperceptible to humans can cause a DNN to label the image as something else entirely (e.g. mislabeling a lion a library). Here we show a related result: it is easy to produce images that are completely unrecognizable to humans, but that state-of-the-art DNNs believe to be recognizable objects with 99.99\% confidence (e.g. labeling with certainty that white noise static is a lion). Specifically, we take convolutional neural networks trained to perform well on either the ImageNet or MNIST datasets and then find images with evolutionary algorithms or gradient ascent that DNNs label with high confidence as belonging to each dataset class. It is possible to produce images totally unrecognizable to human eyes that DNNs believe with near certainty are familiar objects, which we call ``fooling images'' (more generally, fooling examples). Our results shed light on interesting differences between human vision and current DNNs, and raise questions about the generality of DNN computer vision.
\end{abstract}

\section{Introduction}

Deep neural networks (DNNs) learn hierarchical layers of representation from sensory input in order to perform pattern recognition \cite{bengio2009learning, hinton2007learning}.
Recently, these deep architectures have demonstrated impressive, state-of-the-art, and sometimes human-competitive results on many pattern recognition tasks, especially vision classification problems \cite{krizhevsky2012imagenet, dahl2012context, taigman2014deepface, le2011learning}. 
Given the near-human ability of DNNs to classify visual objects, questions arise as to what differences remain between computer and human vision. 

\begin{figure}[!t]
\centering
\includegraphics[width=1.0\columnwidth]{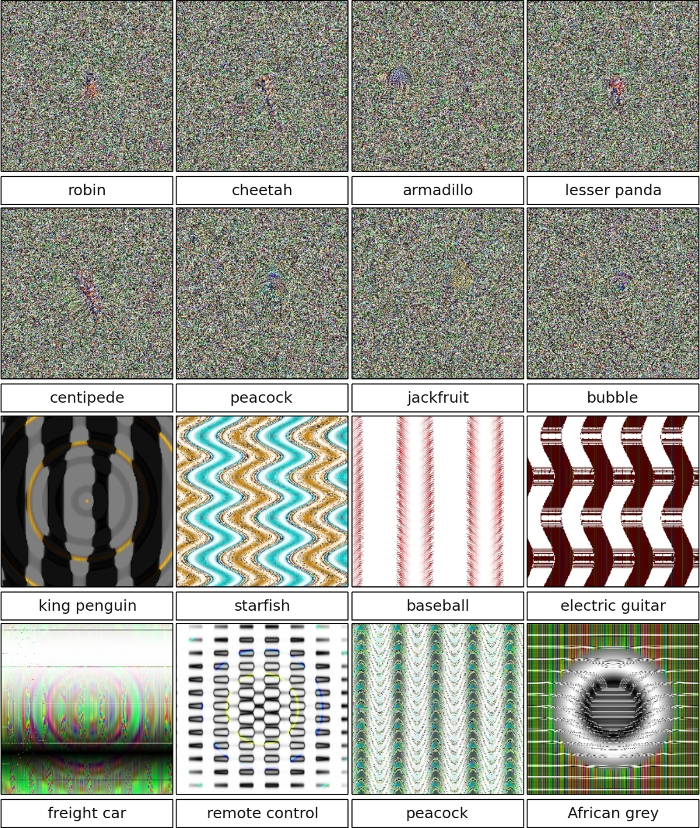}
\caption{
Evolved images that are unrecognizable to humans, but that state-of-the-art DNNs trained on ImageNet believe with $\geq~99.6\%$ certainty to be a familiar object. This result highlights differences between how DNNs and humans recognize objects. 
Images are either directly (\emph{top}) or indirectly (\emph{bottom}) encoded.
}
\label{fig1:DirectAndCPPNfoolignImages}
\end{figure}

A recent study revealed a major difference between DNN and human vision~\cite{szegedy2013intriguing}. Changing an image, originally correctly classified (e.g. as a lion), in a way imperceptible to human eyes, can cause a DNN to label the image as something else entirely (e.g. mislabeling a lion a library). 

\begin{figure*}
\centering
\includegraphics[width=1.8\columnwidth]{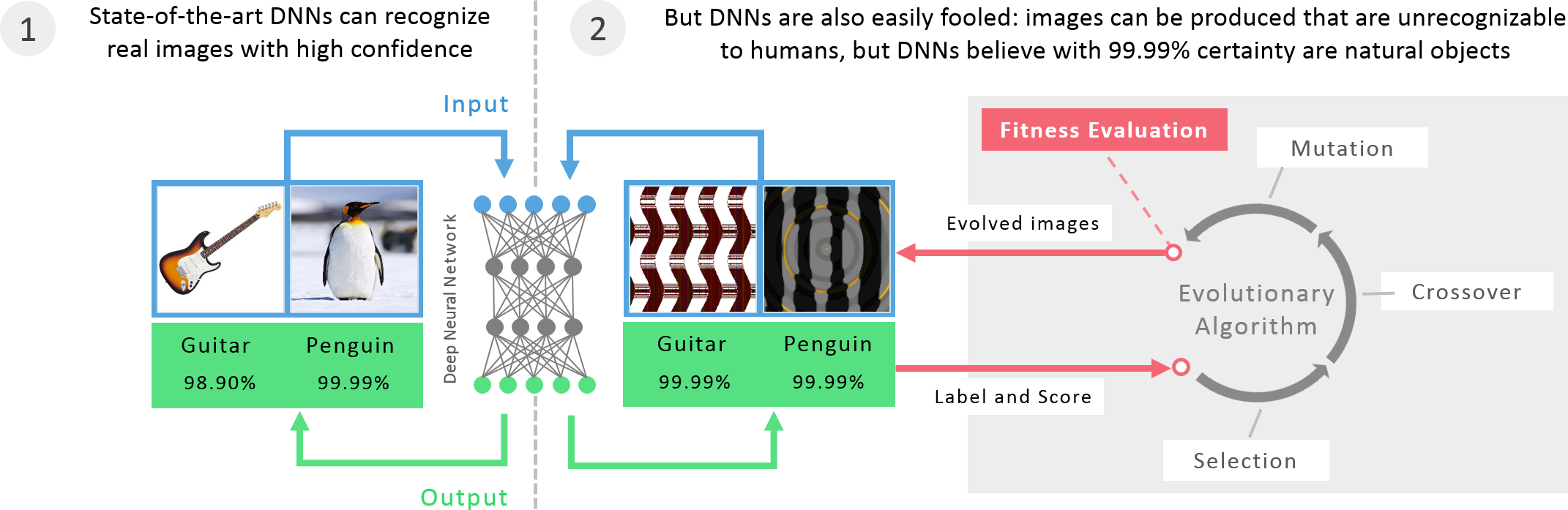}
\caption{
Although state-of-the-art deep neural networks can increasingly recognize natural images (\emph{left panel}), they also are easily fooled into declaring with near-certainty that unrecognizable images are familiar objects (\emph{center}). Images that fool DNNs are produced by evolutionary algorithms (\emph{right panel}) that optimize images to generate high-confidence DNN predictions for each class in the dataset the DNN is trained on (here, ImageNet). 
}
\label{fig:main_concept}
\end{figure*}

In this paper, we show another way that DNN and human vision differ: It is easy to produce images that are completely unrecognizable to humans (Fig.~\ref{fig1:DirectAndCPPNfoolignImages}), but that state-of-the-art DNNs believe to be recognizable objects with over 99\% confidence (e.g. labeling with certainty that TV static is a motorcycle). Specifically, we use evolutionary algorithms or gradient ascent to generate images that are given high prediction scores by convolutional neural networks (convnets) \cite{krizhevsky2012imagenet, lecun1998gradient}. These DNN models have been shown to perform well on both the ImageNet \cite{deng2009imagenet} and MNIST \cite{lecun1998mnist} datasets. 
We also find that, for MNIST DNNs, it is not easy to prevent the DNNs from being fooled by retraining them with fooling images labeled as such. While retrained DNNs learn to classify  the negative examples as fooling images, a new batch of fooling images can be produced that fool these new networks, even after many retraining iterations. 


Our findings shed light on current differences between human vision and DNN-based computer vision. They also raise questions about how DNNs perform in general across different types of images than the ones they have been trained and traditionally tested on. 




\section{Methods}
\label{sec:experiment_framework}

\subsection{Deep neural network models}
To test whether DNNs might give false positives for unrecognizable images, we need a DNN trained to near state-of-the-art performance. We choose the well-known ``AlexNet'' architecture from \cite{krizhevsky2012imagenet}, which is a convnet trained on the 1.3-million-image ILSVRC 2012 ImageNet dataset \cite{deng2009imagenet, russakovsky2014imagenet}. Specifically, we use the already-trained AlexNet DNN  provided by the Caffe software package \cite{jia2014caffe}. It obtains 42.6\% top-1 error rate, similar to the 40.7\% reported by Krizhevsky 2012 \cite{krizhevsky2012imagenet}. While the Caffe-provided DNN has some small differences from Krizhevsky 2012 \cite{krizhevsky2012imagenet}, we do not believe our results would be qualitatively changed by small architectural and optimization differences or their resulting small performance improvements. Similarly, while recent papers have improved upon Krizhevsky 2012, those differences are unlikely to change our results. We chose AlexNet because it is widely known and a trained DNN similar to it is publicly available. In this paper, we refer to this model as ``ImageNet DNN''.

To test that our results hold for other DNN architectures and datasets, we also conduct experiments with the Caffe-provided LeNet model \cite{lecun1998gradient} trained on the MNIST dataset \cite{lecun1998mnist}. The Caffe version has a minor difference from the original architecture in \cite{lecun1998gradient} in that its neural activation functions are rectified linear units (ReLUs) \cite{nair2010rectified} instead of sigmoids. This model obtains 0.94\% error rate, similar to the 0.8\% of LeNet-5 \cite{lecun1998gradient}. We refer to this model as ``MNIST DNN''.

\subsection{Generating images with evolution}
\label{sec:evolutionary_algorithms}

The novel images we test DNNs on are produced by evolutionary algorithms (EAs) ~\cite{floreano2008bio}. EAs are optimization algorithms inspired by Darwinian evolution. They contain a population of ``organisms'' (here, images) that alternately face selection (keeping the best) and then random perturbation (mutation and/or crossover). Which organisms are selected depends on the \emph{fitness function}, which in these experiments is the highest prediction value a DNN makes for that image belonging to a class~(Fig.~\ref{fig:main_concept}).

Traditional EAs optimize solutions to perform well on one objective, or on all of a small set of objectives~\cite{floreano2008bio} (e.g. evolving images to match a single ImageNet class). We instead use a new algorithm called the multi-dimensional archive of phenotypic elites MAP-Elites~\cite{cully2014robots}, which enables us to  simultaneously evolve a population that contains individuals that score well on many classes (e.g. all 1000 ImageNet classes). Our results are unaffected by using the more computationally efficient MAP-Elites over single-target evolution (data not shown).
MAP-Elites works by keeping the best individual found so far for each objective. Each iteration, it chooses a random organism from the population, mutates it randomly, and replaces the current champion for any objective if the new individual has higher fitness on that objective. Here, fitness is determined by showing the image to the DNN; if the image generates a higher prediction score for any class than has been seen before, the newly generated individual becomes the champion in the archive for that class. 


We test EAs with two different \emph{encodings}~\cite{stanley2003taxonomy,clune2011performance}, meaning how an image is represented as a genome. The first has a \emph{direct encoding}, which has one grayscale integer for each of $28\times28$ pixels for MNIST, and three integers (H, S, V) for each of $256\times256$ pixels for ImageNet. Each pixel value is initialized with uniform random noise within the $[0, 255]$ range. Those numbers are independently mutated; first by determining which numbers are mutated, via a rate that starts at 0.1 (each number has a 10\% chance of being chosen to be mutated) and drops by half every 1000 generations. The numbers chosen to be mutated are then altered via the polynomial mutation operator~\cite{deb2001multi} with a fixed mutation strength of 15. The second EA has an \emph{indirect encoding}, which is more likely to produce \emph{regular} images, meaning images that contain compressible patterns (e.g. symmetry and repetition)~\cite{lipson2007principles}. Indirectly encoded images tend to be regular because elements in the genome can affect multiple parts of the image~\cite{stanley2007compositional}.  Specifically, the indirect encoding here is a compositional pattern-producing network (CPPN), which can evolve complex, regular images that resemble natural and man-made objects~\cite{secretan2008picbreeder, stanley2007compositional, auerbach2012automated}.

\begin{figure}[tb]
\centering
\includegraphics[width=1.0\columnwidth]{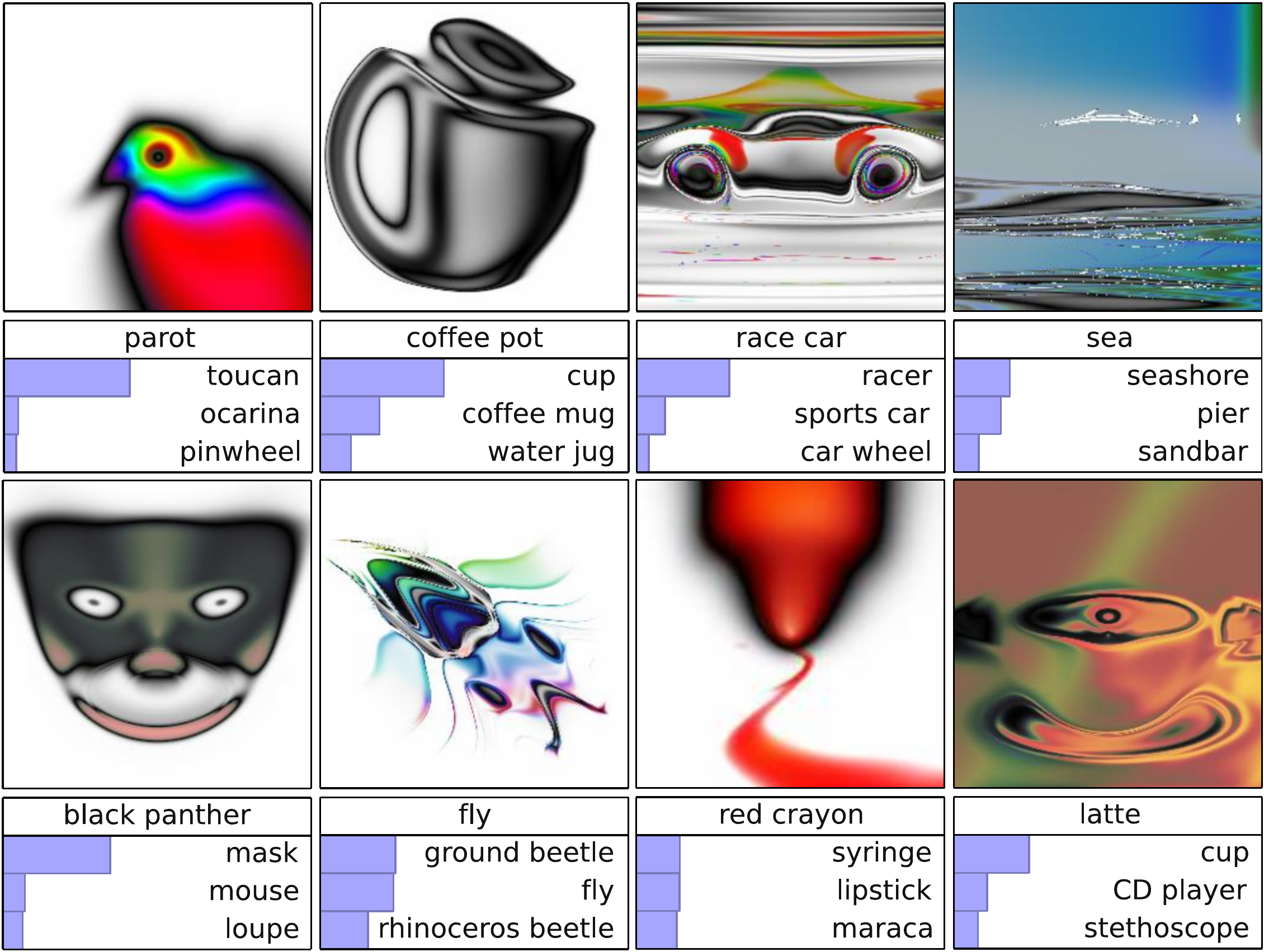}
\caption{ Evolved, CPPN-encoded images produced with humans performing selection on PicBreeder.org. Human image breeders named each object (centered text).  Blue bars show the top three classifications made by a DNN trained on ImageNet (size indicates confidence). Often the first classification relates to the human breeder's label, showing that CPPN-encoded evolution can produce images that humans and DNNs can recognize.
}
\label{fig:picbreeder_8}
\end{figure}

Importantly, images evolved with CPPNs can be recognized by DNNs~(Fig. \ref{fig:picbreeder_8}), providing an existence proof that a CPPN-encoded EA can produce images that both humans and DNNs can recognize. These images were produced on PicBreeder.org~\cite{secretan2008picbreeder}, a site where users serve as the fitness function in an evolutionary algorithm by selecting images they like, which become the parents of the next generation.

CPPNs are similar to artificial neural networks (ANNs). A CPPN takes in the $(x, y)$ position of a pixel as input, and outputs a grayscale value (MNIST) or tuple of HSV color values (ImageNet) for that pixel. Like a neural network, the function the CPPN computes depends on the number of neurons in the CPPN, how they are connected, and the weights between neurons. Each CPPN node can be one of a set of activation functions (here: sine, sigmoid, Gaussian and linear), which can provide geometric regularities to the image. For example, passing the $x$ input into a Gaussian function will provide left-right symmetry, and passing the $y$ input into a sine function provides top-bottom repetition. Evolution determines the topology, weights, and activation functions of each CPPN network in the population.  

As is custom, and was done for the images in Fig. \ref{fig:picbreeder_8}, CPPN networks start with no hidden nodes, and nodes are added over time, encouraging evolution to first search for simple, regular images before adding complexity~\cite{stanley2002evolving}. Our experiments are implemented in the Sferes evolutionary computation framework \cite{mouret2010sferes}.
Our code and parameters are available at \url{http://EvolvingAI.org/fooling}.


%


\section{Results}
\label{sec:fooling_whitenoise}

\subsection{Evolving irregular images to match MNIST}

\begin{figure}[htb]
\centering
\includegraphics[width=1.0\columnwidth]{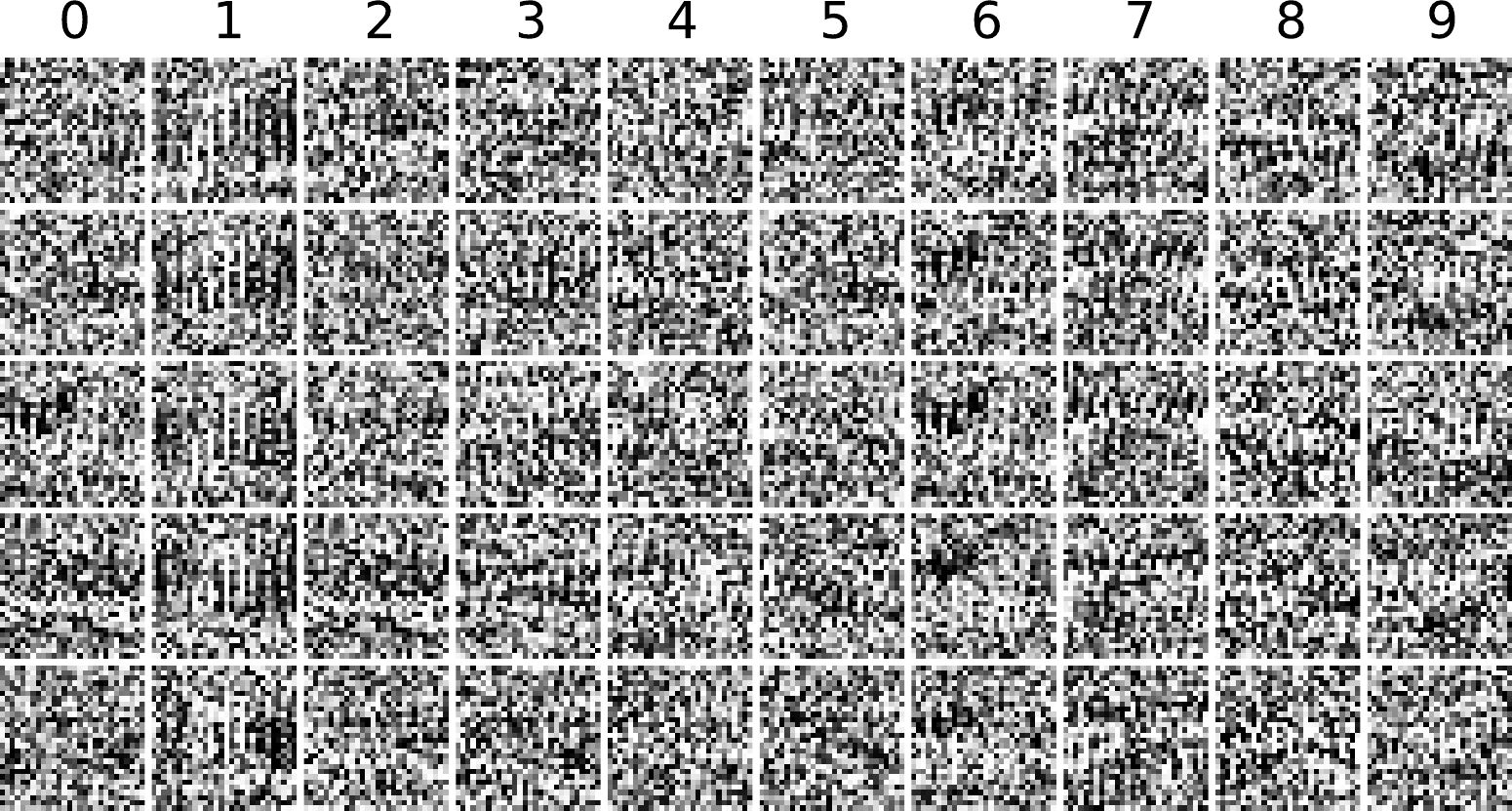}
\caption{Directly encoded, thus irregular, images that MNIST DNNs believe with 99.99\% confidence are digits 0-9. Each column is a digit class, and each row is the result after 200 generations of a randomly selected, independent run of evolution.}
\label{fig:direct_encoding_lenet_mnist_images}
\end{figure}

We first evolve directly encoded images to be confidently declared by LeNet to be digits 0 thru 9 (recall that LeNet is trained to recognize digits from the MNIST dataset).  Multiple, independent runs of evolution repeatedly produce images that MNIST DNNs believe with 99.99\% confidence to be digits, but are unrecognizable as such~(Fig.~\ref{fig:direct_encoding_lenet_mnist_images}). In less than 50 generations, each run of evolution repeatedly produces unrecognizable images of each digit type classified by MNIST DNNs with $\geq 99.99$\% confidence. By 200 generations, median confidence is 99.99\%. Given the DNN's near-certainty, one might expect these images to resemble handwritten digits. On the contrary, the generated images look nothing like the handwritten digits in the MNIST dataset.

\subsection{Evolving regular images to match MNIST}
\label{sec:cppn_mnist_lenet}

\begin{figure}[t]
\centering
\includegraphics[width=1.0\columnwidth]{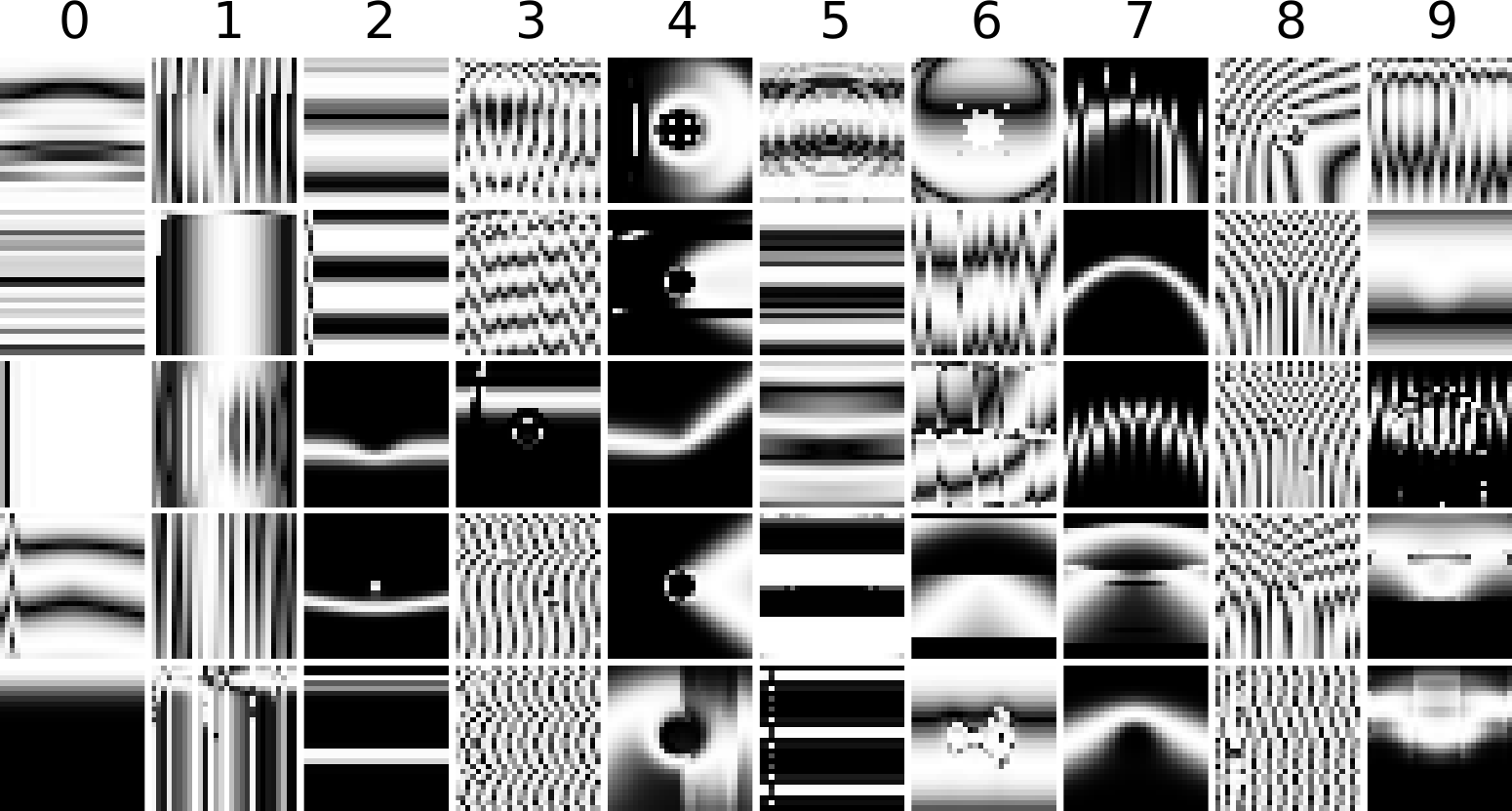}
\caption{Indirectly encoded, thus regular, images that MNIST DNNs believe with 99.99\% confidence are digits 0-9. The column and row descriptions are the same as for Fig.~\ref{fig:direct_encoding_lenet_mnist_images}.
}
\label{fig:mnist_images_5_runs}
\end{figure}

Because CPPN encodings can evolve recognizable images~(Fig.~\ref{fig:picbreeder_8}), we tested whether this more capable, regular encoding might produce more recognizable images than the irregular white-noise static of the direct encoding. The result, while containing more strokes and other regularities, still led to MNIST DNNs labeling unrecognizable images as digits with 99.99\% confidence (Fig.~\ref{fig:mnist_images_5_runs}) after only a few generations. By 200 generations, median confidence is 99.99\%.

Certain patterns repeatedly evolve in some digit classes that appear indicative of that digit~(Fig.~\ref{fig:mnist_images_5_runs}).  Images classified as a 1 tend to have vertical bars, while images classified as a 2 tend to have a horizontal bar in the lower half of the image. Qualitatively similar discriminative features are observed in 50 other runs as well (supplementary material). This result suggests that the EA exploits specific discriminative features corresponding to the handwritten digits learned by MNIST DNNs. 

\subsection{Evolving irregular images to match ImageNet}

We hypothesized that MNIST DNNs might be easily fooled because they are trained on a small dataset that could allow for overfitting (MNIST has only 60,000 training images). To test this hypothesis that a larger dataset might prevent the pathology, we evolved directly encoded images to be classified confidently by a convolutional DNN~\cite{krizhevsky2012imagenet} trained on the ImageNet 2012 dataset, which has 1.3 million natural images in 1000 classes~\cite{deng2012imagenet}. Confidence scores for images were averaged over 10 crops (1 center, 4 corners and 5 mirrors) of size $227\times227$.

\begin{figure}[!h]
\centering
\includegraphics[width=0.87\columnwidth]{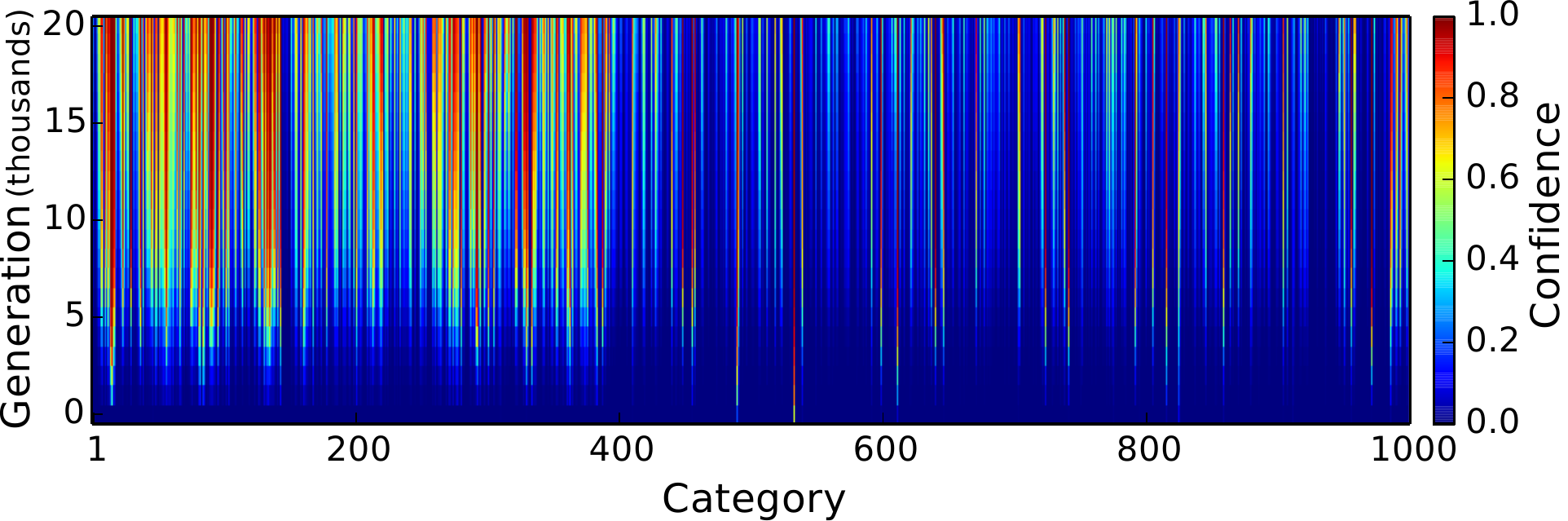}
\caption{Median confidence scores from 5 runs of directly encoded, evolved images for all 1000 ImageNet classes. Though rare, evolution can produce images that the DNN believes with over 99\% confidence to be in a natural, ImageNet class. 
}
\label{fig:direct_encoding_imagenet_heatmap}
\end{figure}

The directly encoded EA was less successful at producing high-confidence images in this case. Even after 20,000 generations, evolution failed to produce high-confidence images for many categories~(Fig.~\ref{fig:direct_encoding_imagenet_heatmap}, median confidence 21.59\%). However, evolution did manage to produce images for 45 classes that are classified with $\geq 99\%$ confidence to be natural images (Fig.~\ref{fig1:DirectAndCPPNfoolignImages}). While in some cases one might discern features of the target class in the image if told the class, humans without such priming would not recognize the image as belonging to that class.

\subsection{Evolving regular images to match ImageNet}
\label{sec:fooling_regular_imagenet_cnn}

Once again, we test whether the CPPN encoding, which has previously evolved images that both humans and DNNs recognize similarly~(Fig.~\ref{fig:picbreeder_8}), might produce more recognizable images than the direct encoding. The hypothesis is that the larger ImageNet dataset and more powerful DNN architecture may interact with the CPPN encoding to finally produce recognizable images. 

\begin{figure}[!h]
\centering
\includegraphics[width=0.87\columnwidth]{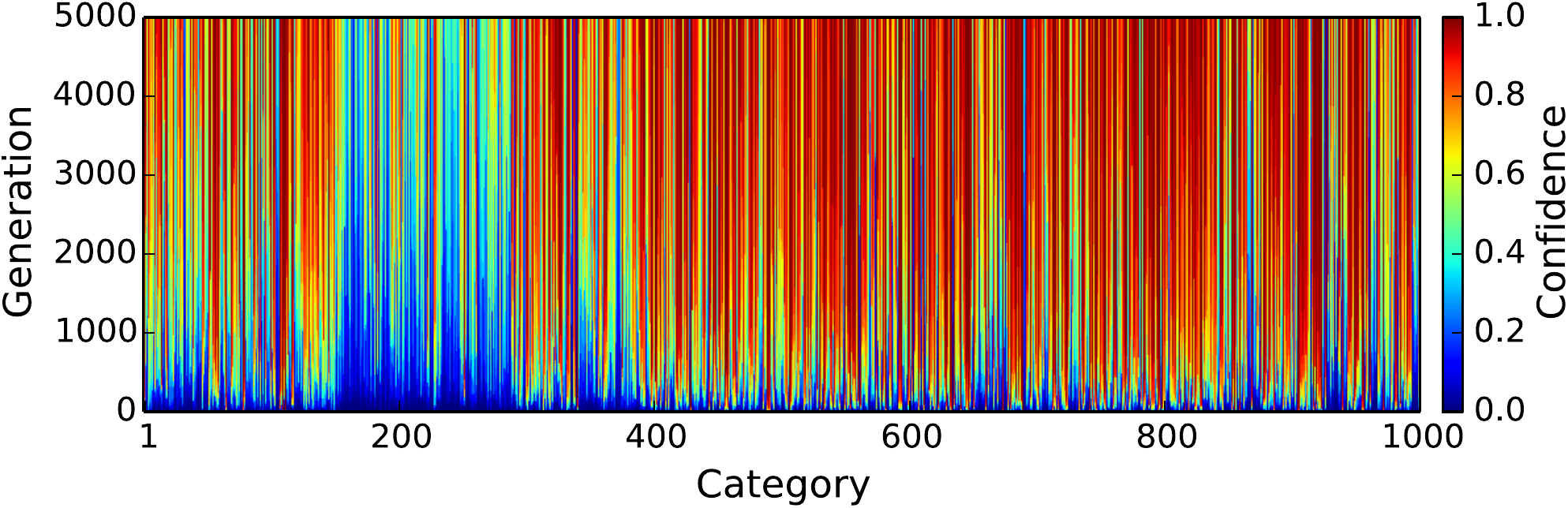}
\caption{Median confidence scores from 5 runs of CPPN-encoded, evolved images for all 1000 ImageNet classes. Evolution can produce many images that the DNN believes with over 99\% confidence to belong to ImageNet classes. 
}
\label{fig:heatmap_5000_gens}
\end{figure}

In five independent runs, evolution produces many images with DNN confidence scores $\geq$ 99.99\%, but that are unrecognizable~(Fig.~\ref{fig1:DirectAndCPPNfoolignImages}~bottom). After 5000 generations, the median confidence score reaches 88.11\%, similar to that for natural images (supplementary material) and significantly higher than the 21.59\% for the direct encoding~(Fig.~\ref{fig:imagenet_vs_imagenet1001}, $p<0.0001$ via Mann-Whitney U test), which was given 4-fold more generations. High-confidence images are found in most categories~(Fig.~\ref{fig:heatmap_5000_gens}). 


\begin{figure}[!h]
\centering
\includegraphics[width=1.0\columnwidth]{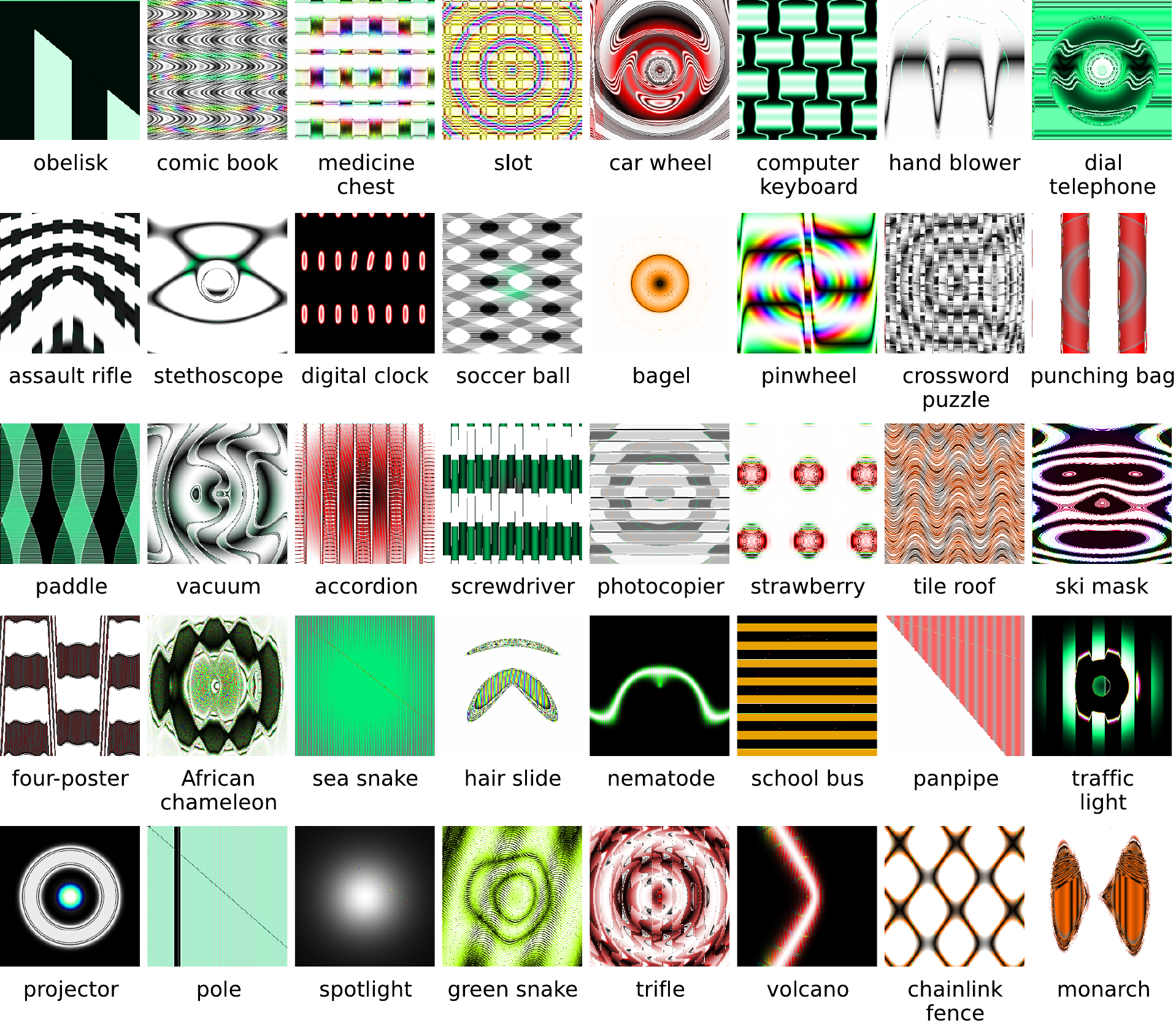}
\caption{Evolving images to match DNN classes produces a tremendous diversity of images. Shown are images selected to showcase diversity from 5 evolutionary runs. The diversity suggests that the images are non-random, but that instead evolutions producing discriminative features of each target class. The mean DNN confidence scores for these images is  $99.12\%$.
}
\label{fig:diversity_40_images}
\end{figure}

While a human not given the class labels for CPPN images would not label them as belonging to that class, the generated images do often contain some features of the target class. For example, in Fig.~\ref{fig1:DirectAndCPPNfoolignImages}, the starfish image contains the blue of water and the orange of a starfish, the baseball has red stitching on a white background, the remote control has a grid of buttons, etc. For many of the produced images, one can begin to identify why the DNN believes the image is of that class once given the class label. This is because evolution need only to produce features that are unique to, or \emph{discriminative} for, a class, rather than produce an image that contains all of the typical features of a class. 

\begin{figure}[!h]
\centering
\includegraphics[width=0.9\columnwidth]{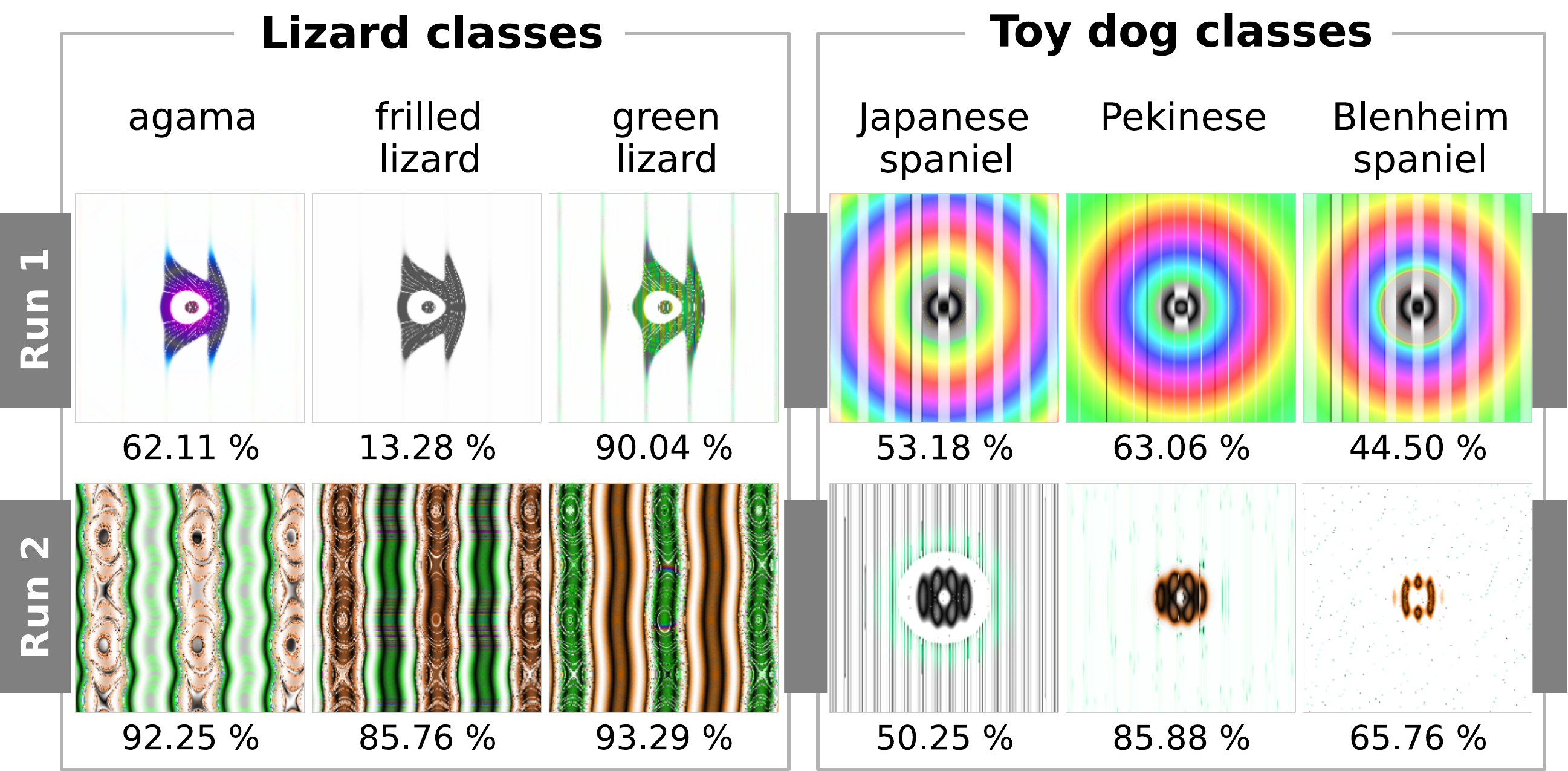}
\caption{Images from the same evolutionary run that fool closely related classes are similar. Shown are the top images evolution generated for three classes that belong to the ``lizard'' parent class, and for three classes that belong to ``toy dog'' parent class. The top and bottom rows show images from independent runs of evolution.
}
\label{fig:branch_similarity}
\end{figure}

The pressure to create these discriminative features led to a surprising amount of diversity in the images produced~(Fig.~\ref{fig:diversity_40_images}).  That diversity is especially noteworthy because (1) it has been shown that imperceptible changes to an image can change a DNN's class label~\cite{szegedy2013intriguing}, so it could have been the case that evolution produced very similar, high-confidence images for all classes, and (2) many of the images are related to each other phylogenetically, which leads evolution to produce similar images for closely related categories~(Fig.~\ref{fig:branch_similarity}). For example, one image type receives high confidence scores for three types of lizards, and a different image type receives high confidence scores for three types of small, fluffy dogs. Different runs of evolution, however, produce different image types for these related categories, revealing that there are different discriminative features per class that evolution exploits. That suggests that there are many different ways to fool the same DNN for each class.    

\begin{figure}[!h]
\centering
\includegraphics[width=0.5\columnwidth]{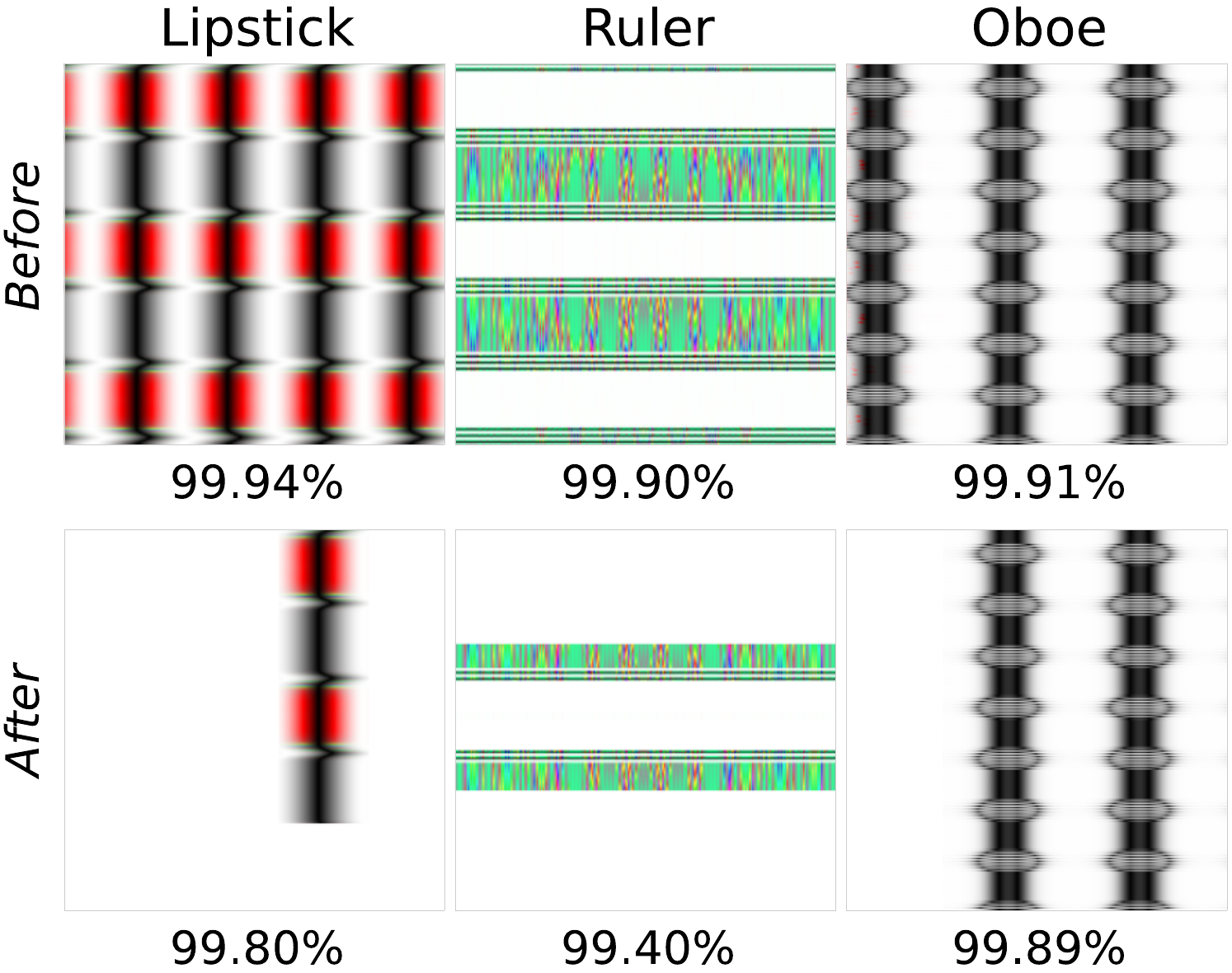}
\caption{\emph{Before}: CPPN-encoded images with repeated patterns. \emph{After}: Manually removing repeated elements suggests that such repetition increases confidence scores.
}
\label{fig:background_removal}
\end{figure}

Many of the CPPN images feature a pattern repeated many times. To test whether that repetition improves the confidence score a DNN gives an image, or whether the repetition stems solely from the fact that CPPNs tend to produce regular images~\cite{stanley2007compositional,clune2011performance}, we ablated (i.e. removed) some of the repeated elements to see if the DNN confidence score for that image drops. Psychologists use the same ablation technique to learn which image features humans use to recognize objects \cite{biederman1995visual}. In many images, ablating extra copies of the repeated element did lead to a performance drop, albeit a small one~(Fig~\ref{fig:background_removal}), meaning that the extra copies make the DNN more confident that the image belongs to the target class. This result is in line with a previous paper \cite{simonyan2013deep} that produced images to maximize DNN confidence scores (discussed below in Section~\ref{karen}), which also saw the emergence of features (e.g. a fox's ears) repeated throughout an image. These results suggest that 
DNNs tend to learn low- and middle-level features rather than the global structure of objects. If DNNs were properly learning global structure, images should receive lower DNN confidence scores if they contain repetitions of object subcomponents that rarely appear in natural images, such as many pairs of fox ears or endless remote buttons~(Fig.~\ref{fig1:DirectAndCPPNfoolignImages}). 

The low-performing band of classes in Fig.~\ref{fig:heatmap_5000_gens} (class numbers 157-286) are dogs and cats, which are overrepresented in the ImageNet dataset (i.e. there are many more classes of cats than classes of cars). One possible explanation for why images in this band receive low confidence scores is that the network is tuned to identify many specific types of dogs and cats. Therefore, it ends up having more units dedicated to this image type than others. In other words, the size of the dataset of cats and dogs it has been trained on is larger than for other categories, meaning it is less overfit, and thus more difficult to fool. If true, this explanation means that larger datasets are a way to ameliorate the problem of DNNs being easily fooled. An alternate, though not mutually exclusive, explanation is that, because there are more cat and dog classes, the EA had difficulty finding an image that scores high in a specific dog category (e.g. Japanese spaniel), but low in any other related categories (e.g. Blenheim spaniel), which is necessary to produce a high confidence given that the final DNN layer is softmax. This explanation suggests that datasets with more classes can help ameliorate fooling.


\subsection{Images that fool one DNN generalize to others}

The results of the previous section suggest that there are discriminative features of a class of images that DNNs learn and evolution exploits. One question is whether different DNNs learn the same features for each class, or whether each trained DNN learns different discriminative features. One way to shed light on that question is to see if images that fool one DNN also fool another. To test that, we evolved CPPN-encoded images with one DNN ($DNN_{A}$) and then input these images to another DNN ($DNN_{B}$). We tested two cases: (1) $DNN_{A}$ and $DNN_{B}$ have identical architectures and training, and differ only in their randomized initializations; and (2) $DNN_{A}$ and $DNN_{B}$ have different DNN architectures, but are trained on the same dataset. We performed this test for both MNIST and ImageNet DNNs. 

Images were evolved that are given $\geq 99.99\%$ confidence scores by both $DNN_{A}$ and $DNN_{B}$. Thus, some general properties of the DNNs are exploited by the CPPN-encoded EA. However, there are also images specifically fine-tuned to score high on $DNN_{A}$, but not on $DNN_{B}$. See the supplementary material for more detail and data. 

\subsection{Training networks to recognize fooling images}
\label{sec:training_cppn}

One might respond to the result that DNNs are easily fooled by saying that, while DNNs are easily fooled when images are optimized to produce high DNN confidence scores, the problem could be solved by simply changing the training regimen to include negative examples. In other words, a network could be retrained and told that the images that previously fooled it should not be considered members of any of the original classes, but instead should be recognized as a new ``fooling images'' class. 

We tested that hypothesis with CPPN-encoded images on both MNIST and ImageNet DNNs. The process is as follows: We train $DNN_{1}$ on a dataset (e.g. ImageNet), then evolve CPPN images that produce a high confidence score for $DNN_{1}$ for the $n$ classes in the dataset, then we take those images and add them to the dataset in a new class $n+1$; then we train $DNN_{2}$ on this enlarged ``+1'' dataset; (optional) we repeat the process, but put the images that evolved for $DNN_{2}$ in the $n+1$ category (a $n+2$ category is unnecessary because any images that fool a DNN are ``fooling images'' and can thus go in the $n+1$ category). Specifically, to represent different types of images, each iteration we add to this $n+1$ category $m$ images randomly sampled from both the first and last generations of multiple runs of evolution that produce high confidence images for $DNN_{i}$. 
Each evolution run on MNIST or ImageNet produces 20 and 2000 images respectively, with half from the first generation and half from the last. Error-rates for trained $DNN_{i}$ are similar to $DNN_{1}$ (supplementary material). 

 

\subsection{Training MNIST DNNs with fooling images}

\begin{figure}[b]
\centering
\includegraphics[width=1\columnwidth]{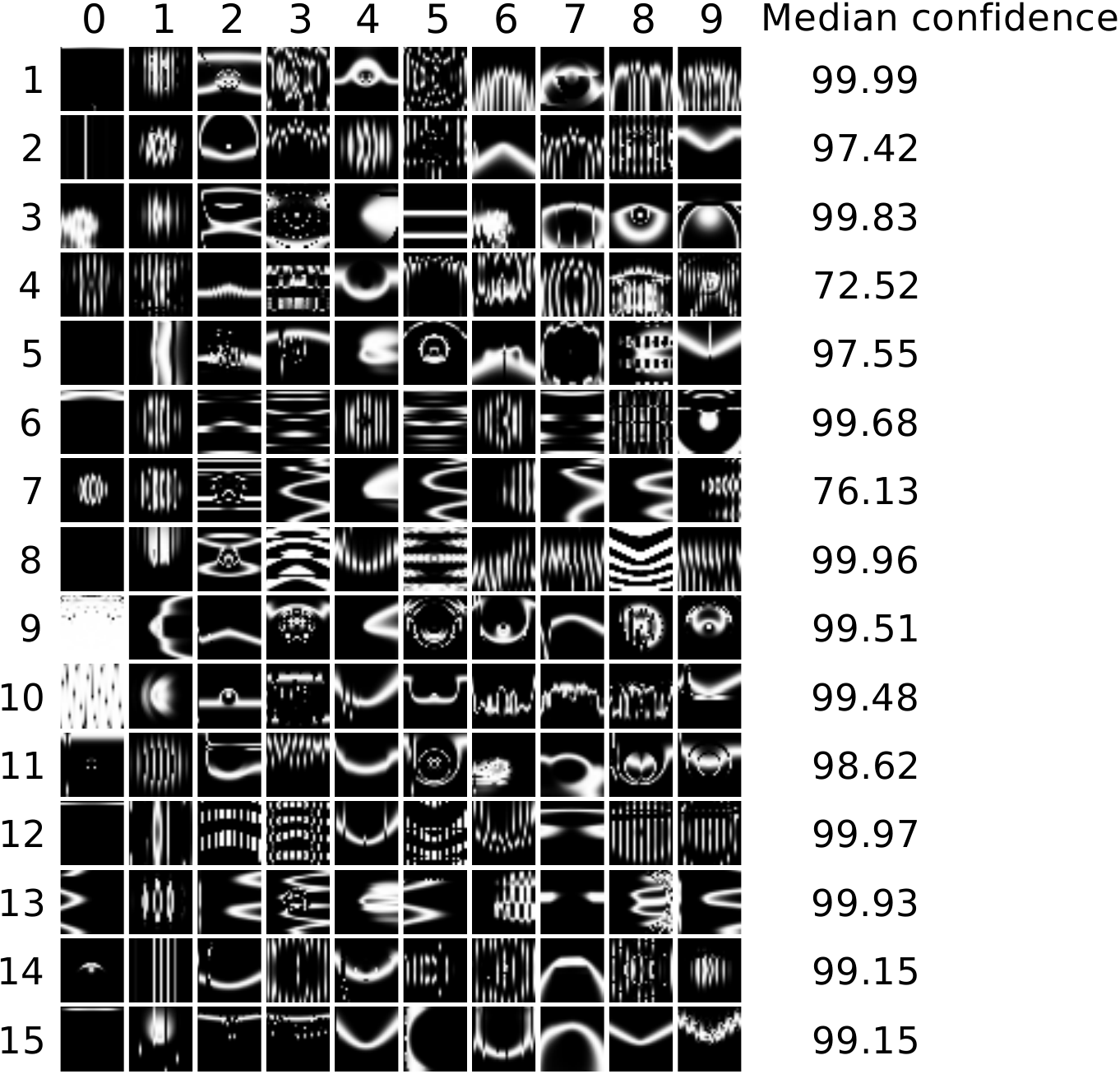}

\caption{Training MNIST $DNN_{i}$ with images that fooled MNIST $DNN_{1}$ through $DNN_{i-1}$ does not prevent evolution from finding new fooling images for $DNN_{i}$. Columns are digits. Rows are $DNN_{i}$ for $i=1...15$. 
Each row shows the 10 final, evolved images from one randomly selected run (of 30) per iteration. Medians are taken from images from all 30 runs.
}
\label{fig:lenet_15_iterations}
\end{figure}

To make the $n+1$ class have the same number of images as other MNIST classes, the first iteration we add 6000 images to the training set (taken from $300$ evolutionary runs). For each additional iteration, we add 1000 new images to the training set. The immunity of LeNet is not boosted by retraining it with fooling images as negative examples. Evolution still produces many unrecognizable images for $DNN_{2}$ with confidence scores of 99.99\%. Moreover, repeating the process for 15 iterations does not help~(Fig.~\ref{fig:lenet_15_iterations}), even though $DNN_{15}$'s overrepresented 11th ``fooling image class'' contains 25\% of the training set images. 


\subsection{Training ImageNet DNNs with fooling images}
\label{sec:training_imagenet_1001}

The original ILSVRC 2012 training dataset was extended with a 1001\textsuperscript{st} class, to which we added 9000 images that fooled $DNN_{1}$. That ~7-fold increase over the ~1300 images per ImageNet class is to emphasize the fooling images in training. Without this imbalance, training with negative examples did not prevent fooling; MNIST retraining did not benefit from over representing the fooling image class.

\begin{figure}[!h]
\centering
\includegraphics[width=0.65\columnwidth]{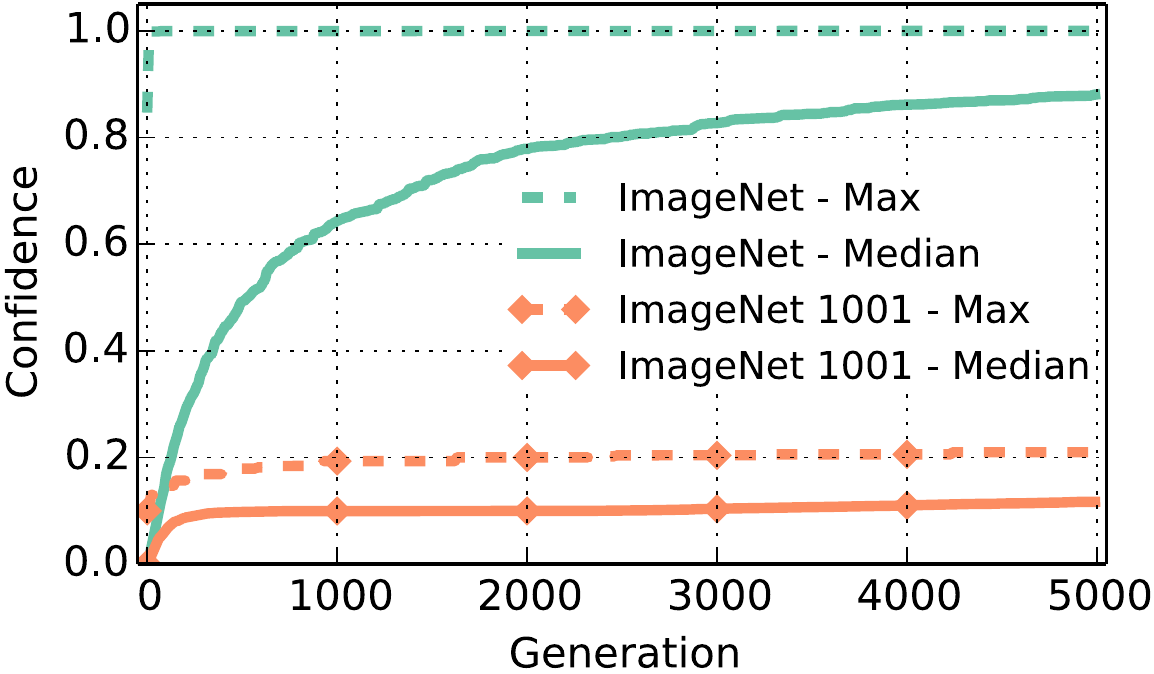}
\caption{Training a new ImageNet DNN ($DNN_{2}$) with images that fooled a previous  DNN ($DNN_{1}$) makes it significantly more difficult for evolution to produce high confidence images. 
}
\label{fig:imagenet_vs_imagenet1001}
\end{figure}

Contrary to the result in the previous section, for ImageNet models, evolution was less able to evolve high confidence images for $DNN_{2}$ than $DNN_{1}$. 
The median confidence score significantly decreased from 88.1\% for $DNN_{1}$ to 11.7\% for $DNN_{2}$ (Fig.~\ref{fig:imagenet_vs_imagenet1001}, $p<0.0001$ via Mann-Whitney U test).
We suspect that ImageNet DNNs were better inoculated against being fooled than MNIST DNNs when trained with negative examples because it is easier to learn to tell CPPN images apart from natural images than it is to tell CPPN images from MNIST digits.

To see whether this $DNN_{2}$ had learned features specific to the CPPN images that fooled $DNN_{1}$, or whether $DNN_{2}$ learned features general to all CPPN images, even recognizable ones, we input recognizable CPPN images from Picbreeder.org to $DNN_{2}$. $DNN_{2}$ correctly labeled 45 of 70 (64\%, top-1 prediction) PicBreeder images as CPPN images, despite having never seen CPPN images like them before. The retrained model thus learned features generic to CPPN images, helping to explain why producing new images that fool $DNN_{2}$ is more difficult.

\subsection{Producing fooling images via gradient ascent}
\label{karen}

\begin{figure}[htb]
\centering
\includegraphics[width=1\columnwidth]{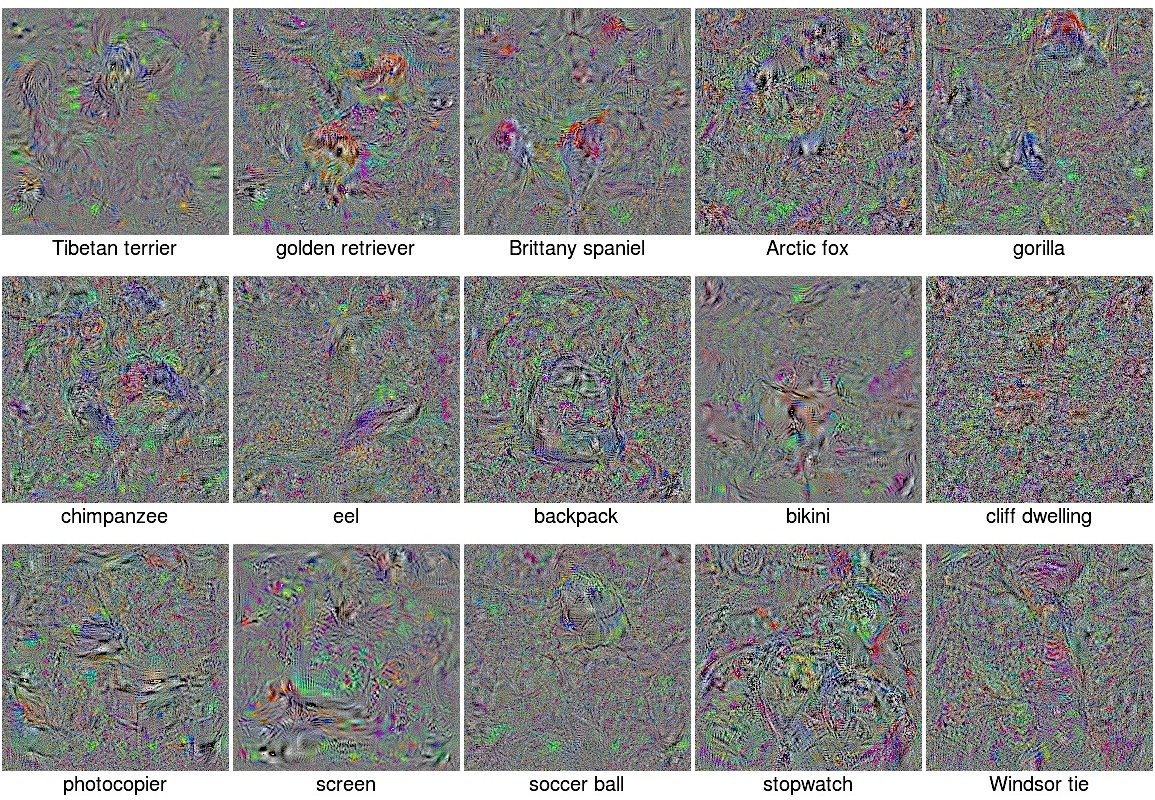}
\caption{Images found by maximizing the softmax output for classes via gradient ascent~\cite{erhan2009visualizing,simonyan2013deep}. Optimization begins at the ImageNet mean (plus small Gaussian noise to break symmetry) and continues until the DNN confidence for the target class reaches 99.99\%. 
Images are shown with the mean subtracted.
Adding regularization makes images more recognizable but results in slightly lower confidence scores (see supplementary material).
}
\label{fig:gradient_descent}
\end{figure}

A different way to produce high confidence, yet mostly unrecognizable images is by using gradient ascent in pixel space \cite{erhan2009visualizing,simonyan2013deep,szegedy2013intriguing}.
We calculate the gradient of the posterior probability for a specific class --- here, a softmax output unit of the DNN --- with respect to the input image using backprop, and then we
follow the gradient to increase a chosen unit's activation.
This technique follows \cite{simonyan2013deep}, but whereas we
aim to find images that produce high confidence classifications, they
sought visually recognizable ``class appearance models.'' By employing L2-regularization,
they produced images with some recognizable features of classes (e.g. dog faces, fox ears, and cup handles).
However, their confidence values are not reported, so to determine the degree to which DNNs are fooled by these backpropagated images, we replicated their work (with some minor changes, see supplementary material) and found that 
images can be made that are also classified by DNNs with 99.99\% confidence, despite them being mostly unrecognizable (Fig.~\ref{fig:gradient_descent}). These optimized images reveal a third method of fooling DNNs that produces qualitatively different examples than the two evolutionary methods in this paper.



\section{Discussion}

\begin{figure}[t]
\centering
\includegraphics[width=1.0\columnwidth]{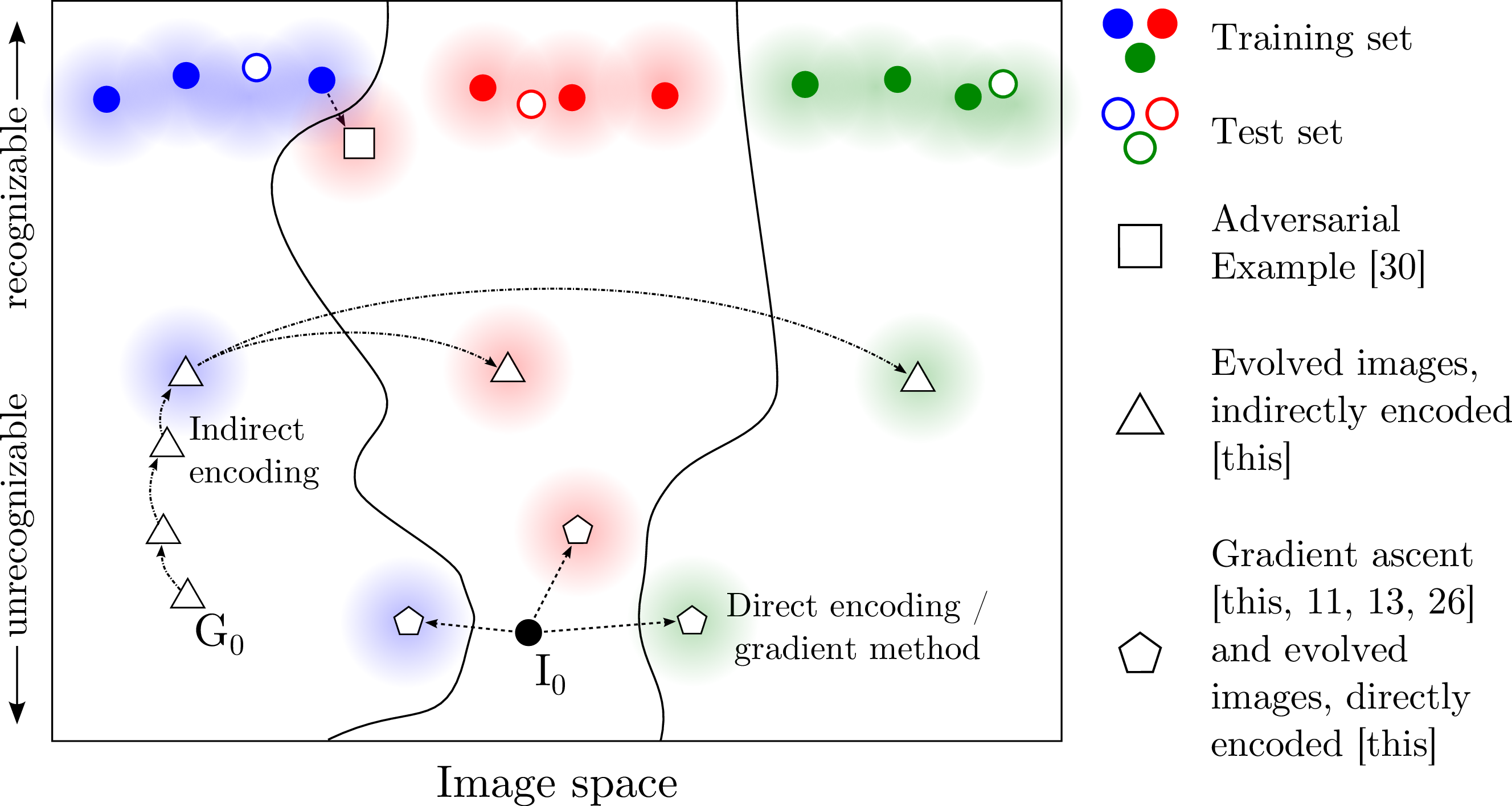}
\caption{
  Interpreting our results and related research. (1) \cite{szegedy2013intriguing} found that an imperceptible change to a correctly classified natural image (blue dot) can result in an image (square) that a DNN classifies as an entirely different class (crossing the decision boundary). The difference between the original image and the modified one is imperceptible to human eyes. (2) It is possible to find high-confidence images (pentagon) using our directly encoded EA or gradient ascent optimization starting from a random or blank image ($I_{0}$) \cite{erhan2009visualizing, goodfellow-2014-arXiv-explaining-and-harnessing-adversarial, simonyan2013deep}. These images have blurry, discriminative features of the represented classes, but do not look like images in the training set. (3) We found that indirectly encoded EAs can find high-confidence, regular images (triangles) that have discriminative features for a class, but are still far from the training set.
}
\label{fig:ann_exploration}
\vspace*{-.8em}
\end{figure}

\begin{figure}[t]
\centering
\includegraphics[width=1.0\columnwidth]{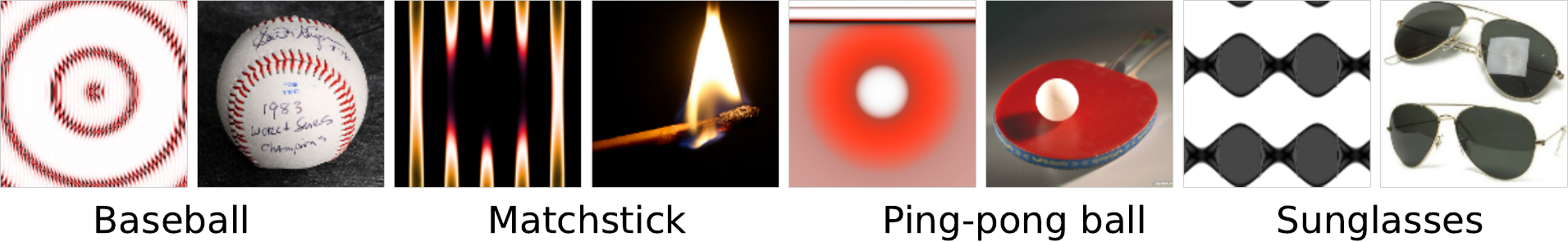}
\caption{Some evolved images do resemble their target class. In each pair, an evolved, CPPN-encoded image (\emph{left}) is shown with a training set image from the target class (\emph{right}).
}
\label{fig:recognizable_images}
\vspace*{-.8em}
\end{figure}

Our experiments could have led to very different results. One might have expected evolution to produce \emph{very similar}, high confidence images for all classes, given that \cite{szegedy2013intriguing} recently showed that imperceptible changes to an image can cause a DNN to switch from classifying it as class A to class B~(Fig.~\ref{fig:ann_exploration}). Instead, evolution produced a tremendous diversity of images~(Figs. \ref{fig1:DirectAndCPPNfoolignImages}, \ref{fig:diversity_40_images}, \ref{fig:background_removal}, \ref{fig:recognizable_images}). Alternately, one might have predicted that evolution would produce \emph{recognizable} images for each class given that, at least with the CPPN encoding, recognizable images have been evolved~(Fig.~\ref{fig:picbreeder_8}). We note that we did not set out to produce unrecognizable images that fool DNNs. Instead, we had hoped the resultant images would be recognizable. 
A different prediction could have been that evolution would fail to produce high confidence scores at all because of local optima. 
It could also have been the case that unrecognizable images would have been given mostly low confidences across all classes instead of a very high confidence for one class.

In fact, none of these outcomes resulted. Instead, evolution produced high-confidence, yet unrecognizable images. Why? Our leading hypothesis centers around the difference between \emph{discriminative models} and \emph{generative models}. 
Discriminative models --- or models that learn $p(y|X)$ for a label vector $y$ and input example $X$ --- like the models in this study, create decision boundaries that partition data into classification regions.
In a high-dimensional input space, the area a discriminative model allocates to a class may be much larger than the area occupied by training examples for that class (see lower 80\% of Fig.~\ref{fig:ann_exploration}).
Synthetic images far from the decision boundary and deep into a classification region may produce high confidence predictions even though they are far from the natural images in the class.
This perspective is confirmed and further investigated by a related study \cite{goodfellow-2014-arXiv-explaining-and-harnessing-adversarial} that shows large regions of high confidence exist in certain discriminative models due to a combination of their locally linear nature and high-dimensional input space.




In contrast, a generative model that represents the complete joint density $p(y,X)$ would enable computing not only $p(y|X)$, but also $p(X)$. Such models may be more difficult to fool because fooling images could be recognized by their low marginal probability $p(X)$, and the DNN's confidence in a label prediction for such images could be discounted when $p(X)$ is low.
Unfortunately, current generative models do not scale well \cite{bengio_icml_2014_deep-generative-stochastic} to the high-dimensionality of datasets like ImageNet, so testing to what extent they may be fooled must wait for advances in generative models.

In this paper we focus on the fact that there exist images that DNNs declare with near-certainty to be of a class, but are unrecognizable as such. However, it is also interesting that some generated images are recognizable as members of their target class once the class label is known. Fig.~\ref{fig:recognizable_images} juxtaposes examples with natural images from the target class. Other examples include the chain-link fence, computer keyboard, digital clock, bagel, strawberry, ski mask, spotlight, and monarch butterfly of Fig.~\ref{fig:diversity_40_images}. To test whether these images might be accepted as art, we submitted them to a selective art competition at the University of Wyoming Art Museum, where they were accepted and displayed (supplementary material).
A companion paper explores how these successes suggest combining DNNs with evolutionary algorithms to make open-ended, creative search algorithms~\cite{nguyen2015innovation}.

The CPPN EA presented can also be considered a novel technique to visualize the features learned by DNNs. The diversity of patterns generated for the same class over different runs (Fig.~\ref{fig:branch_similarity}) indicates the diversity of features learned for that class. Such feature-visualization tools help researchers understand what DNNs have learned and whether features can be transferred to other tasks \cite{yosinski-2014-NIPS-how-transferable-are-features-in-deep}.


One interesting implication of the fact that DNNs are easily fooled is that such false positives could be exploited wherever DNNs are deployed for recognizing images or other types of data. For example, one can imagine a security camera that relies on face or voice recognition being compromised. Swapping white-noise for a face, fingerprints, or a voice might be especially pernicious since other humans nearby might not recognize that someone is attempting to compromise the system. Another area of concern could be image-based search engine rankings: background patterns that a visitor does not notice could fool a DNN-driven search engine into thinking a page is about an altogether different topic. The fact that DNNs are increasingly used in a wide variety of industries, including safety-critical ones such as driverless cars, raises the possibility of costly exploits via techniques that generate fooling images.

\section{Conclusion}

We have demonstrated that discriminative DNN models are easily fooled in that they classify many unrecognizable images with near-certainty as members of a recognizable class. Two different ways of encoding evolutionary algorithms produce two qualitatively different types of unrecognizable ``fooling images'', and gradient ascent produces a third. That DNNs see these objects as near-perfect examples of recognizable images sheds light on remaining differences between the way DNNs and humans recognize objects, raising questions about the true generalization capabilities of DNNs and the potential for costly exploits of solutions that use DNNs. 



\section*{Acknowledgments}

The authors would like to thank Hod Lipson for helpful discussions and the NASA Space Technology Research Fellowship (JY) for funding. We also thank Joost Huizinga, Christopher Stanton, and Jingyu Li for helpful feedback.

{\small
\bibliographystyle{ieee}
\bibliography{egbib}

\begin{thebibliography}{10}\itemsep=-1pt

\bibitem{auerbach2012automated}
J.~E. Auerbach.
\newblock Automated evolution of interesting images.
\newblock In {\em Artificial Life 13}, number EPFL-CONF-191282. MIT Press,
  2012.

\bibitem{bengio2009learning}
Y.~Bengio.
\newblock Learning deep architectures for ai.
\newblock {\em Foundations and trends{\textregistered} in Machine Learning},
  2(1):1--127, 2009.

\bibitem{bengio_icml_2014_deep-generative-stochastic}
Y.~Bengio, E.~Thibodeau-Laufer, G.~Alain, and J.~Yosinski.
\newblock Deep generative stochastic networks trainable by backprop.
\newblock In {\em Proceedings of the 30th International Conference on Machine
  Learning}, 2014.

\bibitem{biederman1995visual}
I.~Biederman.
\newblock {\em Visual object recognition}, volume~2.
\newblock MIT press Cambridge, 1995.

\bibitem{clune2011performance}
J.~Clune, K.~Stanley, R.~Pennock, and C.~Ofria.
\newblock On the performance of indirect encoding across the continuum of
  regularity.
\newblock {\em IEEE Transactions on Evolutionary Computation}, 15(4):346--367,
  2011.

\bibitem{cully2014robots}
A.~Cully, J.~Clune, and J.-B. Mouret.
\newblock Robots that can adapt like natural animals.
\newblock {\em arXiv preprint arXiv:1407.3501}, 2014.

\bibitem{dahl2012context}
G.~E. Dahl, D.~Yu, L.~Deng, and A.~Acero.
\newblock Context-dependent pre-trained deep neural networks for
  large-vocabulary speech recognition.
\newblock {\em Audio, Speech, and Language Processing, IEEE Transactions on},
  20(1):30--42, 2012.

\bibitem{deb2001multi}
K.~Deb.
\newblock {\em Multi-objective optimization using evolutionary algorithms},
  volume~16.
\newblock John Wiley \& Sons, 2001.

\bibitem{deng2012imagenet}
J.~Deng, A.~Berg, S.~Satheesh, H.~Su, A.~Khosla, and L.~Fei-Fei.
\newblock Imagenet large scale visual recognition competition 2012
  (ilsvrc2012), 2012.

\bibitem{deng2009imagenet}
J.~Deng, W.~Dong, R.~Socher, L.-J. Li, K.~Li, and L.~Fei-Fei.
\newblock Imagenet: A large-scale hierarchical image database.
\newblock In {\em Computer Vision and Pattern Recognition, 2009. CVPR 2009.
  IEEE Conference on}, pages 248--255. IEEE, 2009.

\bibitem{erhan2009visualizing}
D.~Erhan, Y.~Bengio, A.~Courville, and P.~Vincent.
\newblock Visualizing higher-layer features of a deep network.
\newblock {\em Dept. IRO, Universit{\'e} de Montr{\'e}al, Tech. Rep}, 2009.

\bibitem{floreano2008bio}
D.~Floreano and C.~Mattiussi.
\newblock {\em Bio-inspired artificial intelligence: theories, methods, and
  technologies}.
\newblock MIT press, 2008.

\bibitem{goodfellow-2014-arXiv-explaining-and-harnessing-adversarial}
I.~J. Goodfellow, J.~Shlens, and C.~Szegedy.
\newblock Explaining and harnessing adversarial examples.
\newblock {\em arXiv preprint arXiv:1412.6572}, Dec. 2014.

\bibitem{hinton2007learning}
G.~E. Hinton.
\newblock Learning multiple layers of representation.
\newblock {\em Trends in cognitive sciences}, 11(10):428--434, 2007.

\bibitem{jia2014caffe}
Y.~Jia, E.~Shelhamer, J.~Donahue, S.~Karayev, J.~Long, R.~Girshick,
  S.~Guadarrama, and T.~Darrell.
\newblock Caffe: Convolutional architecture for fast feature embedding.
\newblock {\em arXiv preprint arXiv:1408.5093}, 2014.

\bibitem{krizhevsky2012imagenet}
A.~Krizhevsky, I.~Sutskever, and G.~E. Hinton.
\newblock Imagenet classification with deep convolutional neural networks.
\newblock In {\em Advances in neural information processing systems}, pages
  1097--1105, 2012.

\bibitem{le2011learning}
Q.~V. Le, W.~Y. Zou, S.~Y. Yeung, and A.~Y. Ng.
\newblock Learning hierarchical invariant spatio-temporal features for action
  recognition with independent subspace analysis.
\newblock In {\em Computer Vision and Pattern Recognition (CVPR), 2011 IEEE
  Conference on}, pages 3361--3368. IEEE, 2011.

\bibitem{lecun1998gradient}
Y.~LeCun, L.~Bottou, Y.~Bengio, and P.~Haffner.
\newblock Gradient-based learning applied to document recognition.
\newblock {\em Proceedings of the IEEE}, 86(11):2278--2324, 1998.

\bibitem{lecun1998mnist}
Y.~LeCun and C.~Cortes.
\newblock The mnist database of handwritten digits, 1998.

\bibitem{lipson2007principles}
H.~Lipson.
\newblock {Principles of modularity, regularity, and hierarchy for scalable
  systems}.
\newblock {\em Journal of Biological Physics and Chemistry}, 7(4):125, 2007.

\bibitem{mouret2010sferes}
J.-B. Mouret and S.~Doncieux.
\newblock Sferes v2: Evolvin'in the multi-core world.
\newblock In {\em Evolutionary Computation (CEC), 2010 IEEE Congress on}, pages
  4079--4086. IEEE, 2010.

\bibitem{nair2010rectified}
V.~Nair and G.~E. Hinton.
\newblock Rectified linear units improve restricted boltzmann machines.
\newblock In {\em Proceedings of the 27th International Conference on Machine
  Learning (ICML-10)}, pages 807--814, 2010.

\bibitem{nguyen2015innovation}
A.~Nguyen, J.~Yosinski, and J.~Clune.
\newblock Introducing the innovation engine: Automated creativity and improved
  stochastic optimization via deep learning.
\newblock In {\em Proceedings of the Genetic and Evolutionary Computation
  Conference}, 2015.

\bibitem{russakovsky2014imagenet}
O.~Russakovsky, J.~Deng, H.~Su, J.~Krause, S.~Satheesh, S.~Ma, Z.~Huang,
  A.~Karpathy, A.~Khosla, M.~Bernstein, et~al.
\newblock Imagenet large scale visual recognition challenge.
\newblock {\em arXiv preprint arXiv:1409.0575}, 2014.

\bibitem{secretan2008picbreeder}
J.~Secretan, N.~Beato, D.~B. D~Ambrosio, A.~Rodriguez, A.~Campbell, and K.~O.
  Stanley.
\newblock Picbreeder: evolving pictures collaboratively online.
\newblock In {\em Proceedings of the SIGCHI Conference on Human Factors in
  Computing Systems}, pages 1759--1768. ACM, 2008.

\bibitem{simonyan2013deep}
K.~Simonyan, A.~Vedaldi, and A.~Zisserman.
\newblock Deep inside convolutional networks: Visualising image classification
  models and saliency maps.
\newblock {\em arXiv preprint arXiv:1312.6034}, 2013.

\bibitem{stanley2002evolving}
K.~Stanley and R.~Miikkulainen.
\newblock Evolving neural networks through augmenting topologies.
\newblock {\em Evolutionary computation}, 10(2):99--127, 2002.

\bibitem{stanley2007compositional}
K.~O. Stanley.
\newblock Compositional pattern producing networks: A novel abstraction of
  development.
\newblock {\em Genetic programming and evolvable machines}, 8(2):131--162,
  2007.

\bibitem{stanley2003taxonomy}
K.~O. Stanley and R.~Miikkulainen.
\newblock A taxonomy for artificial embryogeny.
\newblock {\em Artificial Life}, 9(2):93--130, 2003.

\bibitem{szegedy2013intriguing}
C.~Szegedy, W.~Zaremba, I.~Sutskever, J.~Bruna, D.~Erhan, I.~Goodfellow, and
  R.~Fergus.
\newblock Intriguing properties of neural networks.
\newblock {\em arXiv preprint arXiv:1312.6199}, 2013.

\bibitem{taigman2014deepface}
Y.~Taigman, M.~Yang, M.~Ranzato, and L.~Wolf.
\newblock Deepface: Closing the gap to human-level performance in face
  verification.
\newblock In {\em Computer Vision and Pattern Recognition (CVPR), 2014 IEEE
  Conference on}, pages 1701--1708. IEEE, 2014.

\bibitem{yosinski-2014-NIPS-how-transferable-are-features-in-deep}
J.~{Yosinski}, J.~{Clune}, Y.~{Bengio}, and H.~{Lipson}.
\newblock {How transferable are features in deep neural networks?}
\newblock In Z.~Ghahramani, M.~Welling, C.~Cortes, N.~Lawrence, and
  K.~Weinberger, editors, {\em Advances in Neural Information Processing
  Systems 27}, pages 3320--3328. Curran Associates, Inc., Dec. 2014.

\end{thebibliography}


\begin{thebibliography}{1}\itemsep=-1pt

\bibitem{jia2014caffe}
Y.~Jia, E.~Shelhamer, J.~Donahue, S.~Karayev, J.~Long, R.~Girshick,
  S.~Guadarrama, and T.~Darrell.
\newblock Caffe: Convolutional architecture for fast feature embedding.
\newblock {\em arXiv preprint arXiv:1408.5093}, 2014.

\bibitem{krizhevsky2012imagenet}
A.~Krizhevsky, I.~Sutskever, and G.~E. Hinton.
\newblock Imagenet classification with deep convolutional neural networks.
\newblock In {\em Advances in neural information processing systems}, pages
  1097--1105, 2012.

\bibitem{lecun1998gradient}
Y.~LeCun, L.~Bottou, Y.~Bengio, and P.~Haffner.
\newblock Gradient-based learning applied to document recognition.
\newblock {\em Proceedings of the IEEE}, 86(11):2278--2324, 1998.

\bibitem{russakovsky2014imagenet}
O.~Russakovsky, J.~Deng, H.~Su, J.~Krause, S.~Satheesh, S.~Ma, Z.~Huang,
  A.~Karpathy, A.~Khosla, M.~Bernstein, et~al.
\newblock Imagenet large scale visual recognition challenge.
\newblock {\em arXiv preprint arXiv:1409.0575}, 2014.

\bibitem{szegedy2014going}
C.~Szegedy, W.~Liu, Y.~Jia, P.~Sermanet, S.~Reed, D.~Anguelov, D.~Erhan,
  V.~Vanhoucke, and A.~Rabinovich.
\newblock Going deeper with convolutions.
\newblock {\em arXiv preprint arXiv:1409.4842}, 2014.

\end{thebibliography}
}

\end{document}


\beginsupplementary

\title{Supplementary Material for\\Deep Neural Networks are Easily Fooled:\\High Confidence Predictions for Unrecognizable Images}

\renewcommand{\refname}{Supplementary References}
\renewcommand\thesection{\Alph{section}}


\maketitle

\section{Images that fool one DNN generalize to fool other DNNs}

As we wrote in the paper: ``One question is whether different DNNs learn the same features for each class, or whether each trained DNN learns different discriminative features. One way to shed light on that question is to see if images that fool one DNN also fool another. To test that, we evolved CPPN-encoded images with one DNN ($DNN_{A}$) and then input them to another DNN ($DNN_{B}$), where $DNN_{A}$ and $DNN_{B}$ have identical architectures and training, and differ only in their randomized initializations. We performed this test for both MNIST and ImageNet DNNs.'' Here we show the details of this experiment and its results.

\subsection{Generalization across DNNs with the same architecture}

We performed this test with two MNIST~\cite{lecun1998gradient} DNNs (MNIST$_{A}$ and MNIST$_{B}$) and two ImageNet \cite{krizhevsky2012imagenet} DNNs (ImageNet$_{A}$ and ImageNet$_{B}$), where $A$ and $B$ differ only in their random initializations, but have the same architecture. 300 images were produced with each MNIST DNN, and 1000 images were produced with each ImageNet DNN.

Taking images evolved to score high on $DNN_{A}$ and inputting them to $DNN_{B}$ (and vice versa), we find that there are many evolved images that are given the same top-1 prediction label by both $DNN_{A}$ and $DNN_{B}$ (Table~\ref{table:fool_A_fool_B}a). Furthermore, among those images, many are given $\geq 99.99\%$ confidence scores by both $DNN_{A}$ and $DNN_{B}$ (Table~\ref{table:fool_A_fool_B}b). Thus, evolution produces patterns that are generally discriminative of a class to multiple, independently trained DNNs. On the other hand, there are still images labeled differently by $DNN_{A}$ and $DNN_{B}$ (Table~\ref{table:fool_A_fool_B}a). These images are specifically fine-tuned to exploit the original DNN. We also find $\geq$ 92.18\% of the images that are given the same top-1 prediction label by both networks, are given higher confidence score by the original DNN (Table~\ref{table:fool_A_fool_B}c).


\begin{table}[h]
\footnotesize
\centering
\begin{tabular}{r|c|c|c|c|}
\hline
\multicolumn{1}{|r|}{Dataset} & \multicolumn{2}{c|}{ImageNet} & \multicolumn{2}{c|}{MNIST} \\ \hline
\multicolumn{1}{|r|}{} & \begin{tabular}[c]{@{}c@{}}$DNN_{A}$\\on\\$DNN_{B}$\\images\end{tabular} & \begin{tabular}[c]{@{}c@{}}$DNN_{B}$\\ on\\$DNN_{A}$\\images\end{tabular} & \begin{tabular}[c]{@{}c@{}}$DNN_{A}$\\on\\$DNN_{B}$\\images\end{tabular} & \begin{tabular}[c]{@{}c@{}}$DNN_{B}$\\on\\$DNN_{A}$\\images\end{tabular} \\ \hline
\multicolumn{1}{|r|}{\begin{tabular}[c]{@{}r@{}}Top-1 matches\end{tabular}} & 62.8 & 65.9 & 43.3 & 48.7 \\ \hline

\multicolumn{1}{|r|}{\begin{tabular}[c]{@{}r@{}}(a) Average\end{tabular}}
& \multicolumn{2}{c|}{64.4} & \multicolumn{2}{c|}{46.0} \\ \hline
\multicolumn{1}{|r|}{\begin{tabular}[c]{@{}r@{}}Top-1 matches\\scoring 99\%\end{tabular}} & 5.0 & 7.2 & 27.3 & 27.3 \\ \hline

\multicolumn{1}{|r|}{\begin{tabular}[c]{@{}r@{}}(b) Average\end{tabular}}
& \multicolumn{2}{c|}{6.1} & \multicolumn{2}{c|}{27.3} \\ \hline
\multicolumn{1}{|r|}{\begin{tabular}[c]{@{}r@{}}Top-1 matches\\scoring higher\\on original DNN\end{tabular}} & 95.1 & 98.0 & 88.5 & 95.9 \\ \hline

\multicolumn{1}{|r|}{\begin{tabular}[c]{@{}r@{}}(c) Average\end{tabular}}
& \multicolumn{2}{c|}{96.6} & \multicolumn{2}{c|}{92.2} \\ \cline{2-5} \hline
\end{tabular}
\caption{
\small\newline\underline{Top-1 matches}: The percent of images that are given the same top-1 label by both $DNN_{A}$ and $DNN_{B}$.
\newline\underline{Top-1 matches scoring 99\%}: The percent of images for which both $DNN_{A}$ and $DNN_{B}$ believe the top-1 predicted label to be the same and the two confidence scores given are both $\geq$ 99\%. 
\newline\underline{Top-1 matches scoring higher}: Of the images that are given the same top-1 label by both $DNN_{A}$ and $DNN_{B}$, the percent that are given a higher confidence score by the original DNN than by the other, testing DNN.
}
\label{table:fool_A_fool_B}
\end{table}

\begin{figure}[htb]
\centering
\includegraphics[width=1.0\columnwidth]{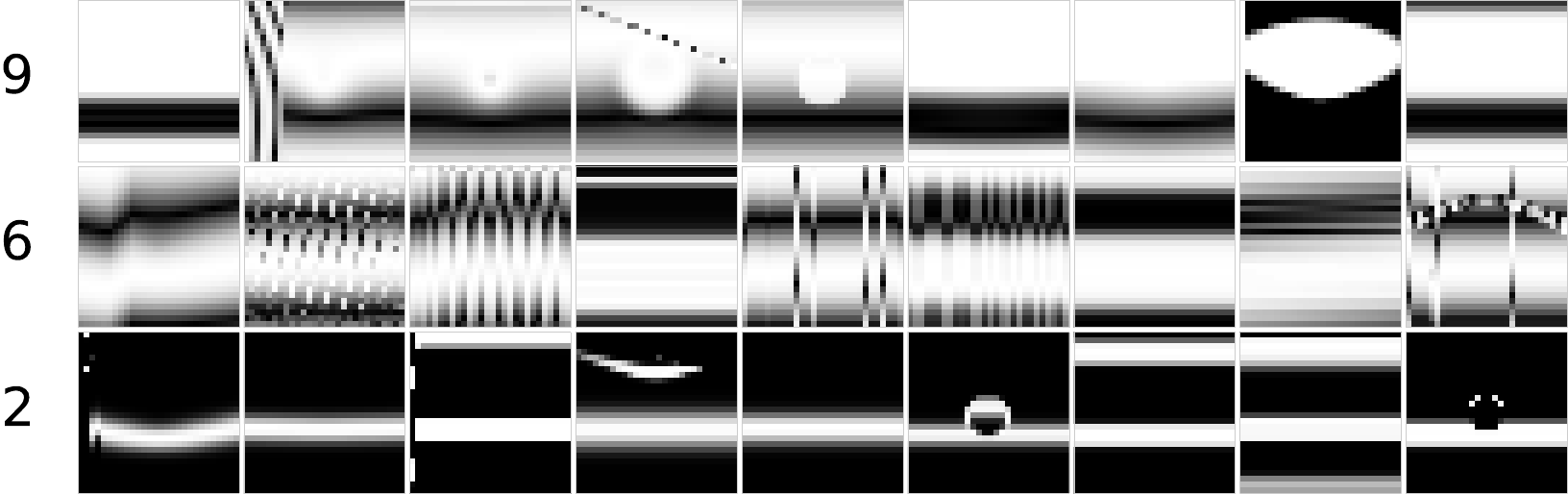}
\caption{CPPN-encoded, evolved images which are given $\geq$ 99\% confidence scores by both $DNN_{A}$ and $DNN_{B}$ to represent digits 9, 6, and 2. 
Each column represents an image produced by an independent run of evolution, yet evolution converges on a similar design, which fools not just the DNN it evolved with, but another, independently trained DNN as well.
}
\label{fig:category_269}
\end{figure}

From the experiment with MNIST DNNs, we observed that images evolved to represent digit classes 9, 6, and 2 fooled both networks $DNN_{A}$ and $DNN_{B}$ the most. Furthermore, these images revealed distinctive patterns (Figure~\ref{fig:category_269}).

\subsection{Generalization across DNNs that have different architectures}

Here we test whether images that fool a DNN with one architecture also fool another DNN with a  different architecture. We performed this test with two well-known ImageNet DNN architectures: AlexNet \cite{krizhevsky2012imagenet} and GoogLeNet \cite{szegedy2014going}, both of which are provided by Caffe \cite{jia2014caffe} and trained on the same ILSVRC 2012 dataset \cite{russakovsky2014imagenet}. GoogLeNet has a top-1 error rate of 31.3\%.

1000 images were produced with each ImageNet DNN. We found that
20.7\% of images evolved for GoogLeNet are also given the same top-1 label by AlexNet (and 17.3\% vice versa). Thus, many fooling examples are not fit precisely to a particular network, but generalize across different DNN architectures.

\section{Does using an ensemble of networks instead of just one prevent fooling?}

We also tested whether requiring an image to fool an ensemble of multiple networks makes it impossible to produce fooling images. We tested an extreme case where each network in the ensemble has a different architecture. Specifically, we tested with an ensemble of 3 different DNN architectures: CaffeNet, AlexNet and GoogLeNet. CaffeNet \cite{jia2014caffe} performs similarly to AlexNet \cite{krizhevsky2012imagenet}, but has a slightly different architecture. The final confidence score given to an image is calculated as the mean of the three scores given by these three different DNNs. After only 4000 generations, evolution was still able to produce fooling images for 231 of the 1000 classes with $\geq90$\% confidence. Moreover, the median is also high at 65.2\% and the max is 100\%.

\section{Training networks to recognize fooling images to prevent fooling}

As we wrote in the paper: ``One might respond to the result that DNNs are easily fooled by saying that, while DNNs are easily fooled when images are optimized to produce high DNN confidence scores, the problem could be solved by simply changing the training regimen to include negative examples. In other words, a network could be retrained and told that the images that previously fooled it should not be considered members of any of the original classes, but instead should be recognized as a new fooling images class.'' 

We tested this hypothesis with CPPN-encoded images on both MNIST and ImageNet DNNs. The process is as follows: We train $DNN_{1}$ on a dataset (e.g. ImageNet), then evolve CPPN images that are given a high confidence score by $DNN_{1}$ for the $n$ classes in the dataset, then we take those images and add them to the dataset in a new class $n+1$; then we train $DNN_{2}$ on this enlarged ``+1'' dataset; (optional) we repeat the process, but put the images that evolved for $DNN_{2}$ in the $n+1$ category (a $n+2$ category is unnecessary because any images that fool a DNN are ``fooling images'' and can thus go in the $n+1$ category). 

Specifically, to represent different types of images, each iteration we add to this $n+1$ category $m$ images. These images are randomly sampled from both the first and last generations of multiple runs of evolution that produce high confidence images for $DNN_{i}$. 
Each run of evolution on MNIST or ImageNet produces 20 or 2000 images, respectively, with half from the first generation and half from the last. As in the original experiments evolving images for MNIST, each evolution run on MNIST or ImageNet lasts for 200 or 5000 generations, respectively. These generation numbers were chosen from the previous experiments. The specific training details are presented in the following sections. 

\subsection{Training MNIST DNNs with fooling images}

To make the $n+1$ class have the same number of images as other MNIST classes, the first iteration we add 6000 and 1000 images to the training and validation sets, respectively. For each additional iteration, we add 1000 and 100 new images to the training and validation sets (Table~\ref{table:training_details_mnist}). 

MNIST DNNs ($DNN_{1}-DNN_{15}$) were trained on images of size $28\times28$, using stochastic gradient descent (SGD) with a momentum of 0.9. Each iteration of SGD used a batch size of 64, and a multiplicative weight decay of 0.0005. The learning rate started at 0.01, and reduced every iteration by an \emph{inverse} learning rate policy (defined in Caffe \cite{jia2014caffe}) with power $=0.75$ and gamma $=0.0001$. $DNN_{2}-DNN_{15}$ obtained similar error rates to the 0.94\% of $DNN_{1}$ trained on the original MNIST (Table~\ref{table:training_details_mnist}). 

Since evolution still produced many unrecognizable images for $DNN_{2}$ with confidence scores of 99.99\%, we repeated the process for 15 iterations (Table~\ref{table:training_details_mnist}). However, the retraining does not help, even though $DNN_{15}$'s overrepresented 11th ``fooling image class'' contains $\sim$25\% of the training set images. 

\begin{table}[htb]
\begin{tabular}{cccccc}
\hline
${i}$ & Error & MNIST Error & Train & Val & Score \\ \hline
1         & 0.94 & 0.94          & 60000    & 10000      &99.99 \\
2         & 1.02 & 0.87          & 66000    & 11000      &97.42 \\
3         & 0.92 & 0.87          & 67000    & 11100      &99.83 \\
4         & 0.89 & 0.83          & 68000    & 11200      &72.52 \\
5         & 0.90 & 0.96          & 69000    & 11300      &97.55 \\
6         & 0.89 & 0.99          & 70000    & 11400      &99.68 \\
7         & 0.86 & 0.98          & 71000    & 11500      &76.13 \\
8         & 0.91 & 1.01          & 72000    & 11600      &99.96 \\
9         & 0.90 & 0.86          & 73000    & 11700      &99.51 \\
10        & 0.84 & 0.94          & 74000    & 11800      &99.48 \\
11        & 0.80 & 0.93          & 75000    & 11900      &98.62 \\
12        & 0.82 & 0.98          & 76000    & 12000      &99.97 \\
13        & 0.75 & 0.90          & 77000    & 12100      &99.93 \\
14        & 0.80 & 0.96          & 78000    & 12200      &99.15 \\
15        & 0.79 & 0.95          & 79000    & 12300      &99.15 \\
\end{tabular}
\caption{Details of 15 training iterations of MNIST DNNs. $DNN_{1}$ is the model trained on the original MNIST dataset without CPPN images. $DNN_{2}-DNN_{15}$ are models trained on the extended dataset with CPPN images added.
\small\newline\underline{Error}: The error (\%) on the validation set (with CPPN images added).
\newline\underline{MNIST Error}: The error (\%) on the original MNIST validation set (10,000 images).
\newline\underline{Train}: The number of images in the training set.
\newline\underline{Val}: The number of images in the validation set.
\newline\underline{Score}: The median confidence scores (\%) of images produced by evolution for that iteration. These numbers are also provided in the paper.
}
\label{table:training_details_mnist}
\end{table}

\subsection{Training ImageNet DNNs with fooling images}
The original ILSVRC 2012 training dataset was extended with a 1001\textsuperscript{st} class, to which we added 9000 images and 2000 images that fooled $DNN_{1}$ to the training and validation sets, respectively. That $\sim$7-fold increase over the $\sim$1300 training images per ImageNet class is to emphasize the fooling images in training. Without this imbalance, training with negative examples did not prevent fooling; retrained MNIST DNNs did not benefit from this strategy of over representing the fooling image class (data not shown). 

The images produced by $DNN_{1}$ are of size $256\times256$ but cropped to $227\times227$ for training. $DNN_{2}$ was trained using SGD with a momentum of 0.9. Each iteration of SGD used a batch size of 256, and a multiplicative weight decay of 0.0005. The learning rate started at 0.01, and dropped by a factor of 10 every 100,000 iterations. Training stopped after 450,000 iterations. The whole training procedure took $\sim$10 days on an Nvidia K20 GPU.

Training $DNN_{2}$ on ImageNet yielded a top-1 error rate of 41.0\%, slightly better than the 42.6\% for $DNN_{1}$: we hypothesize the improved error rate is because the 1001\textsuperscript{st} CPPN image class is easier than the other 1000 classes, because it represents a different \emph{style} of images, making it easier to classify them. Supporting this hypothesis is the fact that $DNN_{2}$ obtained a top-1 error rate of 42.6\% when tested on the original ILSVRC 2012 validation set. 

In contrast to the result in the previous section, for ImageNet models, evolution was less able to evolve high confidence images for $DNN_{2}$ compared to the high confidences evolution produced for $DNN_{1}$. The median confidence score significantly decreased from 88.1\% for $DNN_{1}$ to 11.7\% for $DNN_{2}$ ($p<0.0001$ via Mann-Whitney U test).


\section{Evolving regular images to match MNIST}

\begin{figure*}[htb]
\centering
\includegraphics[width=.97\columnwidth]{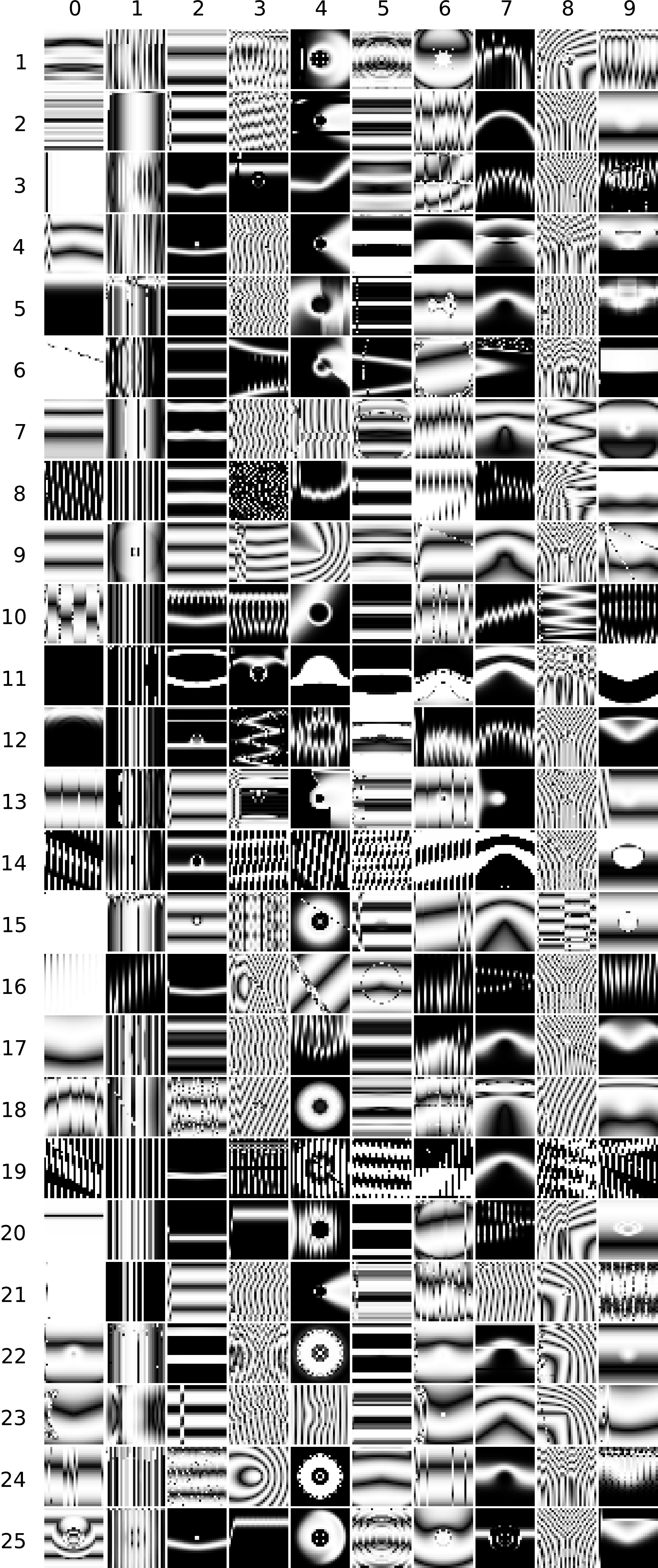}
\hspace{.06\columnwidth}
\includegraphics[width=.97\columnwidth]{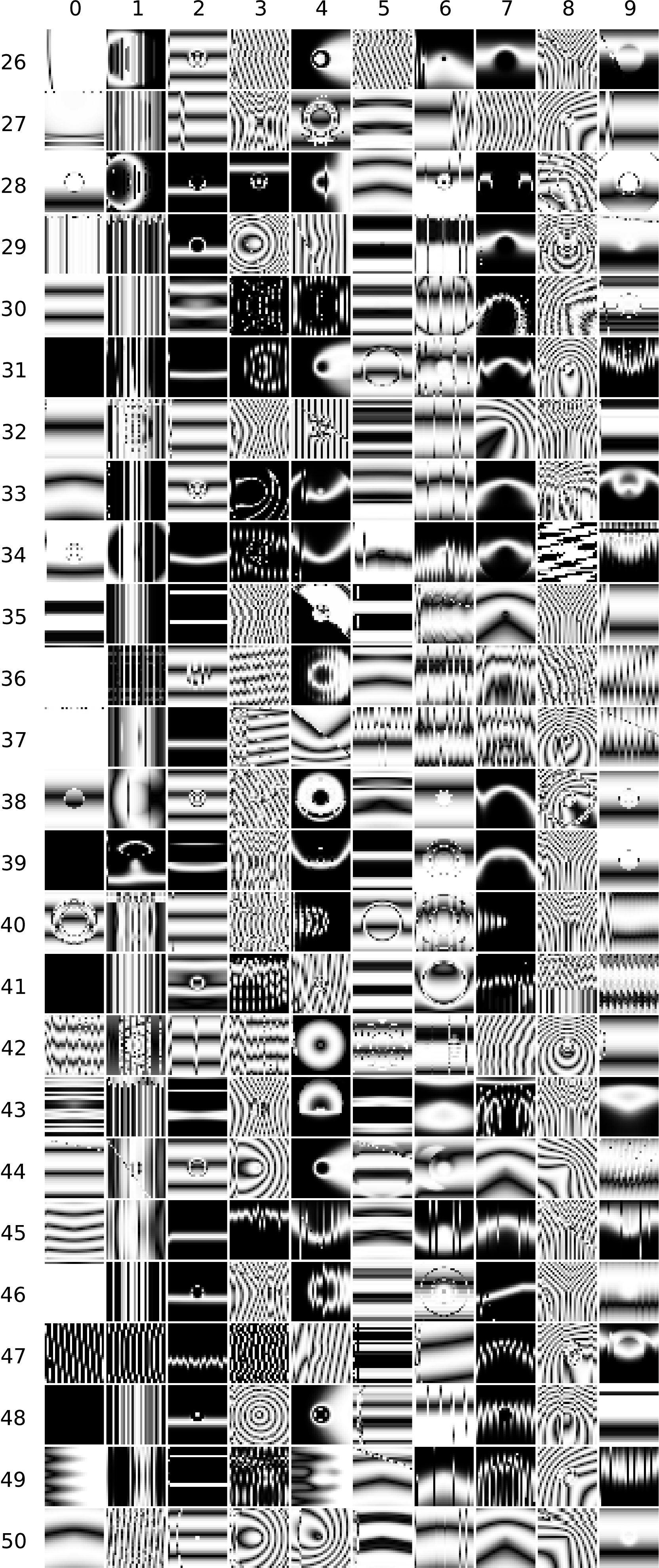}
\caption{
	50 independent runs of evolution produced images that an MNIST DNN believes with 99.99\% to be handwritten digits. Columns are digits. In each row are the final (best) images evolved for each class during that run.
}
\label{fig:indirect_mnist_1_50}
\end{figure*}

As we wrote in the paper: ``Because CPPN encodings can evolve recognizable images, we tested whether this more capable, regular encoding might produce more recognizable images than the irregular white-noise static of the direct encoding. The result, while containing more strokes and other regularities, still led to LeNet labeling unrecognizable images as digits with 99.99\% confidence after only a few generations. By 200 generations, median confidence is 99.99\%.''. Here we show $10$ images $\times50$ runs $= 500$ images produced by the CPPN-encoded EA that an MNIST DNN believes with 99.99\% to be handwritten digits (Fig.~\ref{fig:indirect_mnist_1_50}). 

Looking at these images produced by 50 independent runs of evolution, one can observe that images classified as a 1 tend to have vertical bars. Images classified as a 2 tend to have a horizontal bar in the lower half of the image. Moreover, since an 8 can be drawn by mirroring a 3 horizontally, the DNN may have learned some common features from these two classes from the training set. Evolution repeatedly produces similar patterns for class 3 and class 8. 

\section{Gradient ascent with regularization}

\begin{figure*}[htb]
\centering
\includegraphics[width=1.0\textwidth]{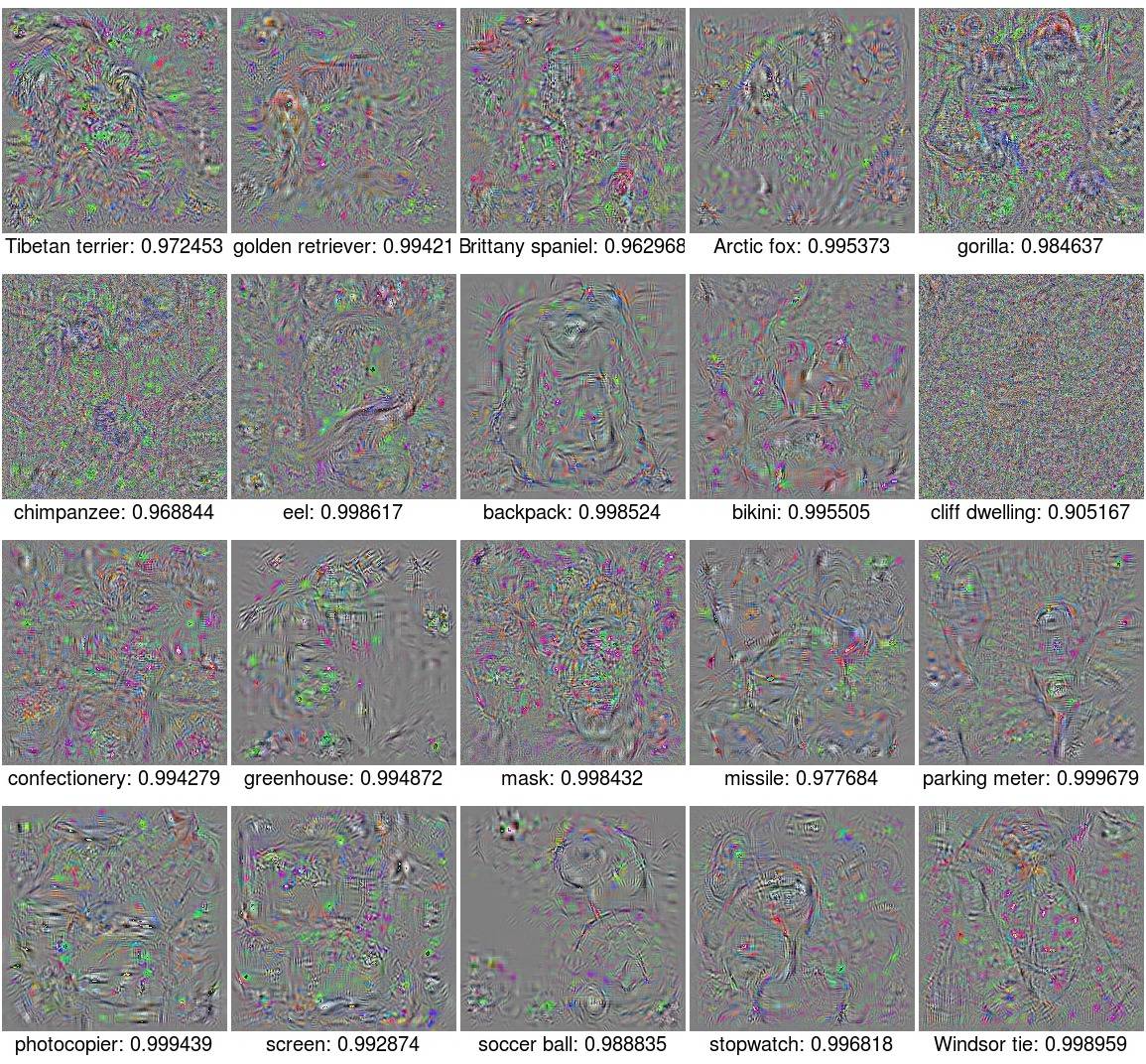}
\caption{Images found by directly maximizing an objective function consisting of the posterior probability (softmax output) added to a regularization term, here L2-regularization. Optimization begins at the ImageNet mean plus small Gaussian noise to break symmetry.
When regularization is added, confidences are generally lower than 99.99\% because the objective contains terms other than confidence. Here, the average is 98.591\%.
For clarity, images are shown with the mean subtracted.}
\label{fig:decay}
\end{figure*}

\begin{figure*}[ht]
\centering
\includegraphics[width=1.0\textwidth]{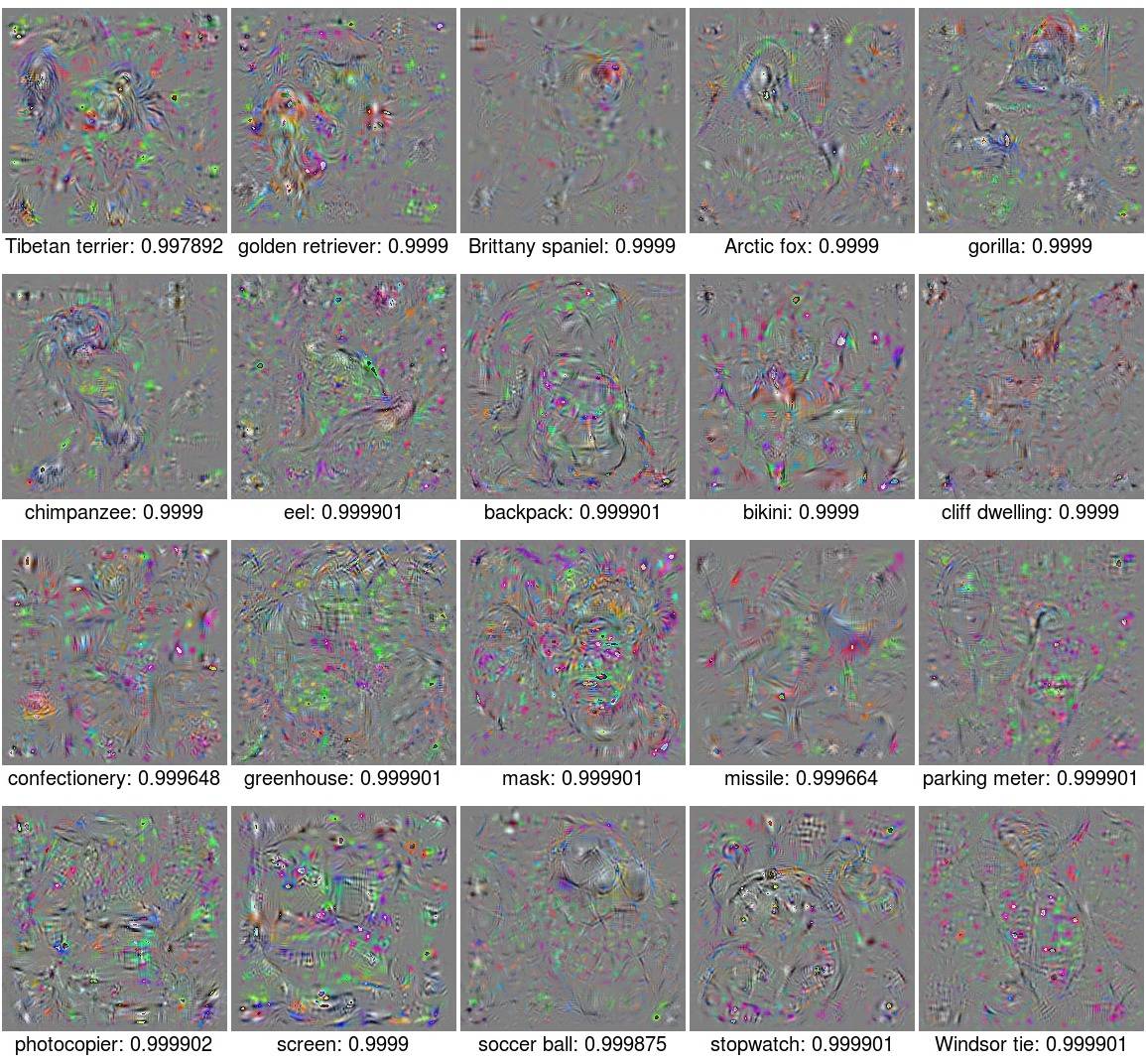}
\caption{As in Fig.~\ref{fig:decay}, but with blurring and pseudo-L1-regularization, which is accomplished by setting the pixels with lowest norm to zero throughout the optimization.}
\label{fig:decay_blur_chop}
\end{figure*}

\begin{figure*}[ht]
\centering
\includegraphics[width=1.0\textwidth]{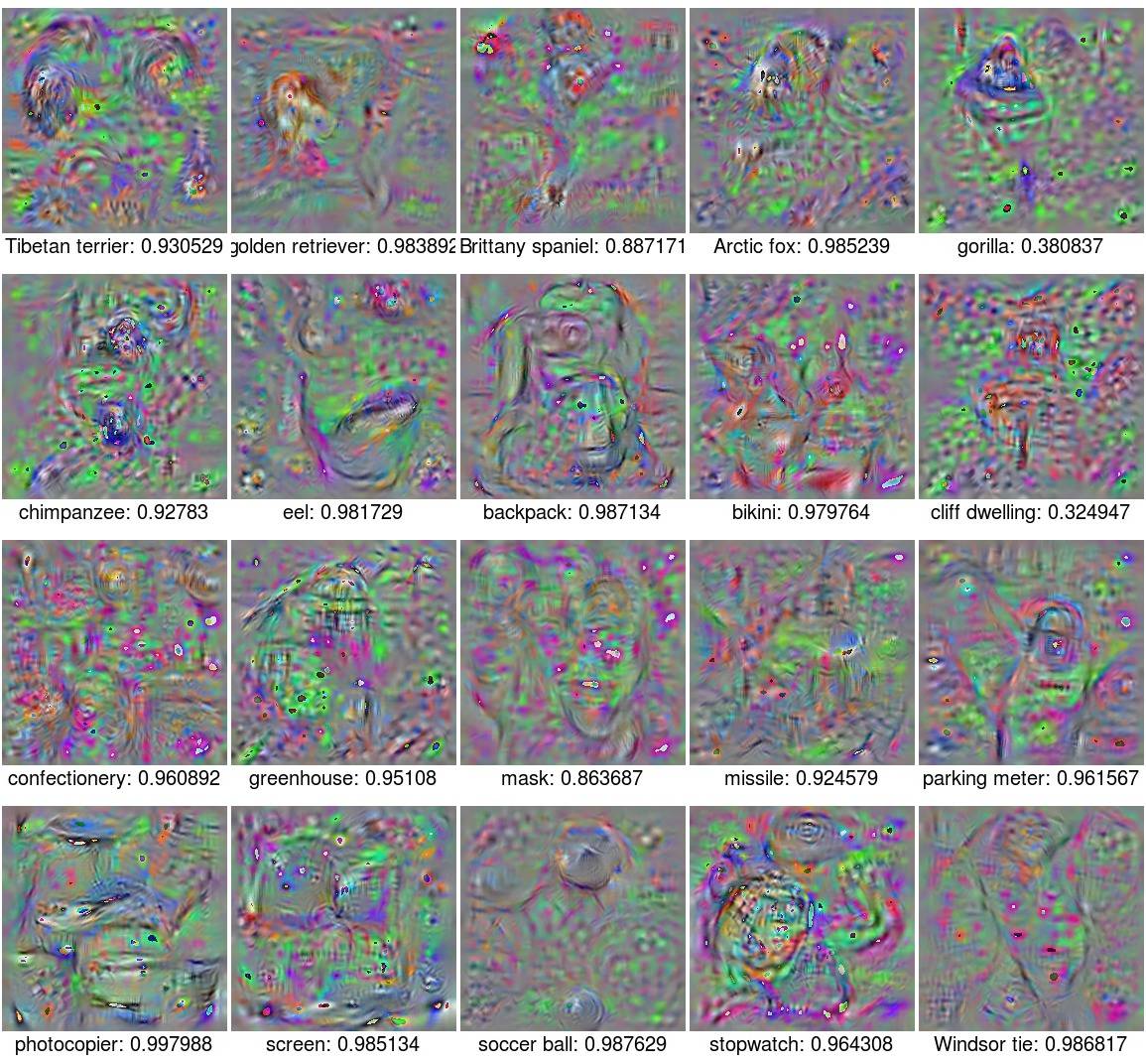}
\caption{As in Fig.~\ref{fig:decay}, but with slightly more aggressive blurring than in Fig.~\ref{fig:decay_blur_chop}.}
\label{fig:aggres_decay_blur}
\end{figure*}

In the paper we showed images produced by direct gradient ascent to
maximize the posterior probability (softmax output) for 20 example
classes. Directly optimizing this objective quickly produces
confidence over $99.99\%$ for unrecognizable images. By adding different
types of regularization, we can also produce more recognizable images.
We tried three types of regularization, highlighted in the Figs.~\ref{fig:decay},~\ref{fig:decay_blur_chop}, and~\ref{fig:aggres_decay_blur}.


Fig.~\ref{fig:decay} shows L2-regularization, implemented as a weight
decay each step. At each step of the optimization, the current
mean-subtracted image $X$ is multiplied by a constant $1-\gamma$ for
small $\gamma$. Fig.~\ref{fig:decay} shows $\gamma = 0.01$.

Fig.~\ref{fig:decay_blur_chop} shows weight decay (now with $\gamma =
0.001$) plus two other types of regularization. The first additional
regularization is a small blurring operator applied each step to bias
the search toward images with less high frequency information and more
low frequency information. This was implemented via a Gaussian blur
with radius $0.3$ after every gradient step. The second additional
regularization was a pseudo-L1-regularization in which the (R, G, B) pixels with
norms lower than the $20^{th}$ percentile were set to 0. This tended to
produce slightly sparser images.

Finally, Fig.~\ref{fig:aggres_decay_blur} shows a lower learning rate
with the same weight decay and slightly more aggressive blurring.
Because the operations of weight decay and blurring do not depend on
the learning rate, this produces an objective containing far more
regularization.  As a result, many of the classes never achieve
$99\%$, but the visualizations are of a different quality and, in
some cases, more clear.

\begin{figure*}[ht]
\centering
\includegraphics[width=1.0\textwidth]{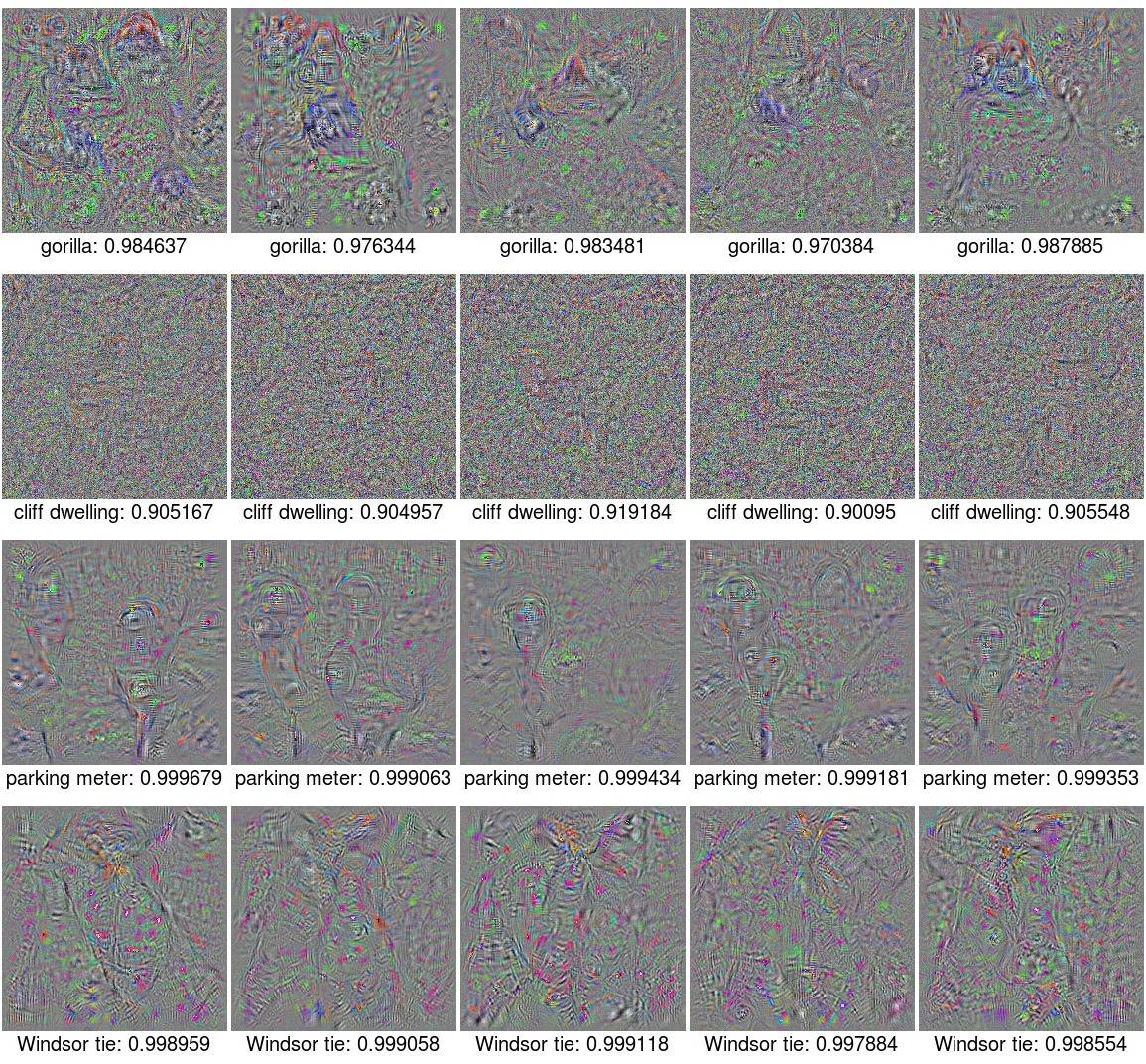}
\caption{Multiple images produced for each class in the manner of Fig.~\ref{fig:decay}. Each column shows the result of a different local optimum, which was reached by starting at the ImageNet mean and adding different draws of small Gaussian noise.}
\label{fig:decay_mult}
\end{figure*}

\begin{figure*}[ht]
\centering
\includegraphics[width=1.0\textwidth]{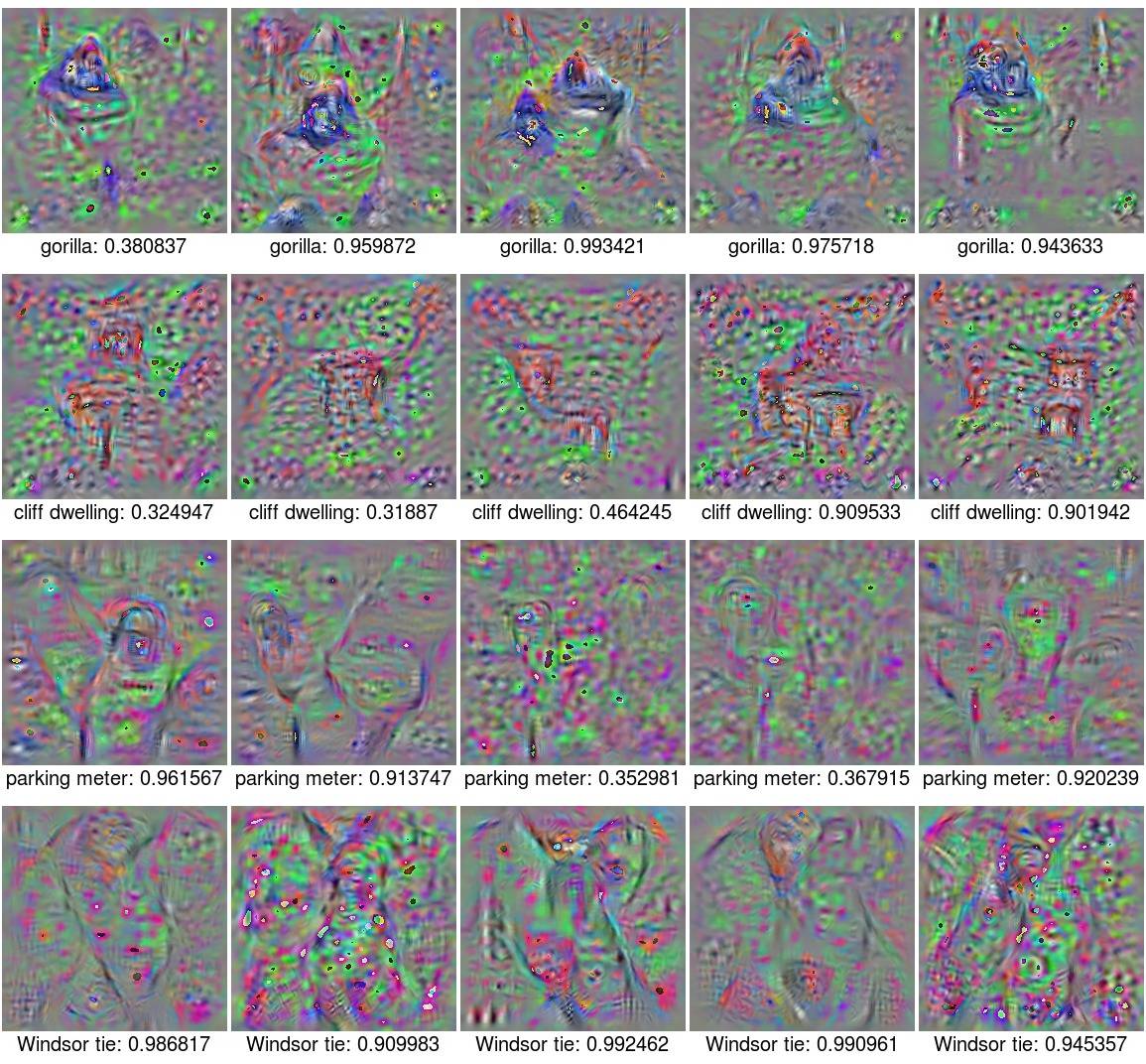}
\caption{Multiple images produced for each class in the manner of Fig.~\ref{fig:aggres_decay_blur}. Each column shows the result of a different local optimum, which was reached by starting at the ImageNet mean and adding different draws of small Gaussian noise.}
\label{fig:aggres_decay_blur_mult}
\end{figure*}

All images generated in this manner are optimized by starting at the
ImageNet mean plus a small amount of Gaussian noise to break symmetry
 and then following
the gradient. The noise has a standard deviation of $1/255$ along each dimension, where dimensions have been scaled to fall into the range $[0,1]$.
Because of this random initialization, the final image
produced depends on the random draw of Gaussian noise.
Fig.~\ref{fig:decay_mult} and
Fig.~\ref{fig:aggres_decay_blur_mult} show the variety of images that
may be produced by taking different random draws of this initial noise.

\section{Confidence scores of real ImageNet images}

The optimization methods presented can generate unrecognizable images that are given high confidence scores. However, to find out if these high scores for fooling images are similar to the confidence scores given by DNNs for the natural images they were trained to classify, we evaluate the entire ImageNet validation set with the ImageNet DNN \cite{krizhevsky2012imagenet}. Across 50,000 validation images, the median confidence score is 60.3\%. Across the cases when images are classified correctly (i.e., the top-1 label matches the ground truth label), the DNN gives a median confidence score of 86.7\%. On the contrary, when the top-1 prediction label does not match the ground truth, the images are given only 33.7\% median confidence. Thus, the median confidence score of 88.11\% of synthetic images that match ImageNet is comparable to that of real images.

\section{Can the fooling images be considered art?}
\begin{figure}[htb]
\centering
\includegraphics[width=1.0\columnwidth]{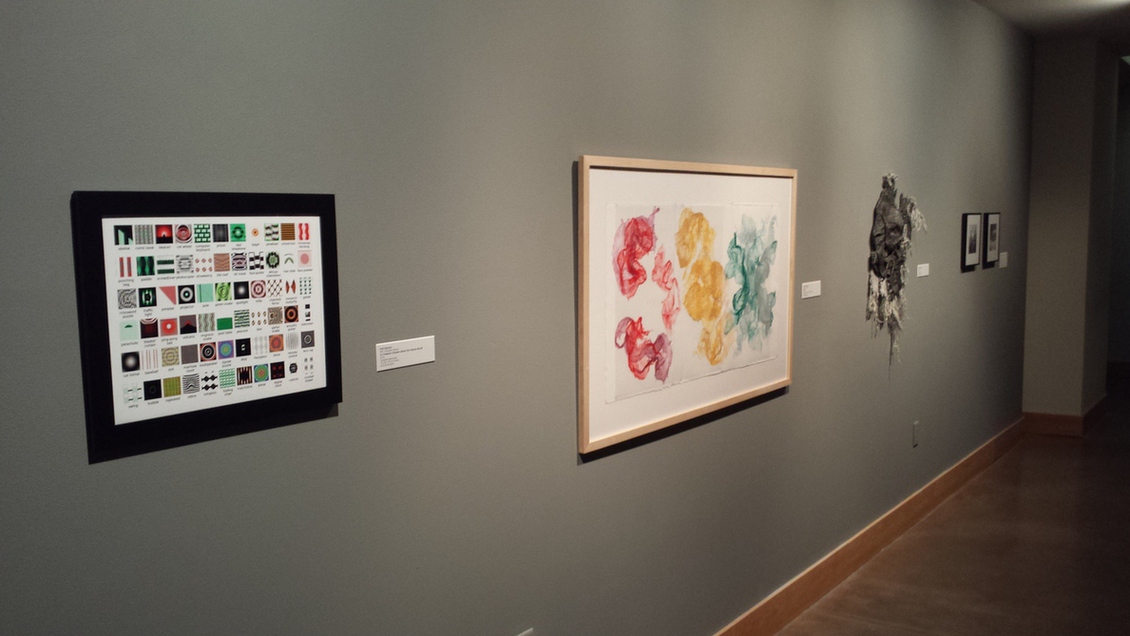}
\caption{A selection of fooling images were accepted as art in a selective art competition. They were then displayed alongside human-made art at a museum.
}
\label{fig:museum_exhibition}
\end{figure}

\begin{figure}[htb]
\centering
\includegraphics[width=1.0\columnwidth]{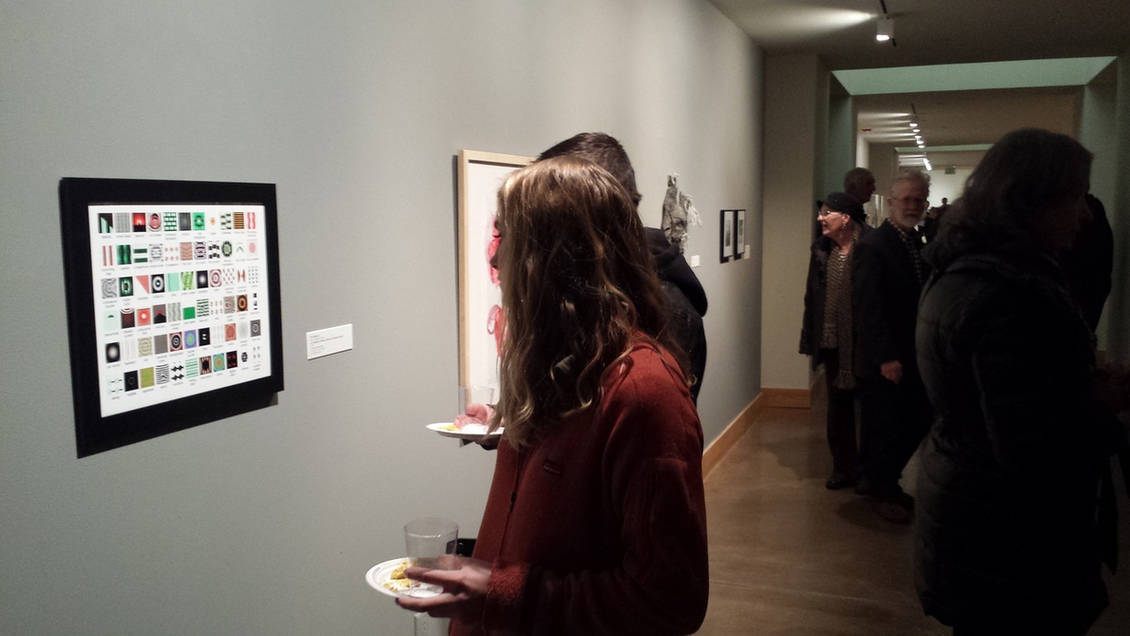}
\caption{Museum visitors view a montage of CPPN-encoded fooling images.
}
\label{fig:museum_people_looking}
\end{figure}

To test the hypothesis that the CPPN fooling images could actually be considered art, we submitted a selection of them to a selective art contest: the ``University of Wyoming 40th Annual Juried Student Exhibition'', which only accepted 35.5\% of the submissions. Not only were the images accepted, but they were also amongst the 21.3\% of submissions to be given an award. The work was then displayed at the University of Wyoming Art Museum (Fig.~\ref{fig:museum_exhibition},~\ref{fig:museum_people_looking}). The submitted image is available at \url{http://evolvingai.org/fooling}.

{\small
\bibliographystyle{ieee}
\bibliography{egbib}
}

\clearpage

%

